\documentclass[11pt]{article}

\usepackage[final]{acl}

\usepackage{times}
\usepackage{latexsym}

\usepackage[T1]{fontenc}

\usepackage[utf8]{inputenc}

\usepackage{microtype}

\usepackage{inconsolata}

\usepackage{graphicx}
\usepackage{amsmath}
\usepackage{booktabs}
\usepackage{enumitem}
\usepackage{multirow}
\usepackage[table]{xcolor}
\usepackage{tabularx}
\usepackage{subcaption}
\usepackage{placeins}
\usepackage{caption}
\title{BoundRL: Efficient Token-level Structured Text Segmentation through Reinforced Boundary Generation}

\author{
Haoyuan Li$_1$\thanks{Work done during an internship at Amazon Web Service.}, Zhengyuan Shen$_2$\thanks{Corresponding Author}, Sullam Jeoung$_2$, Yueyan Chen$_2$ \AND Jiayu Li$_2$, Qi Zhu$_2$ , Shuai Wang$_2$, Vassilis Ioannidis$_2$, Huzefa Rangwala$_2$ \thanks{work done at Amazon} \\
$^1$University of North Carolina at Chapel Hill, $^2$Amazon Web Services \\
\texttt{haoyuanl@cs.unc.edu}; \{\texttt{donshen}, \texttt{sullamij}, \texttt{jlijiayu}, \texttt{qzhuamzn}, \\
\texttt{wshui}, \texttt{ivasilei}, \texttt{rhuzefa}\}\texttt{@amazon.com}
}

\newcommand{\task}{\texttt{BoundRL}}
\newcommand{\dataset}{\texttt{StructSeg}}

\begin{document}
\maketitle
\begin{abstract}
Structured texts refer to texts containing structured elements beyond plain texts, such as code snippets and placeholders. Such structured texts increasingly require segmentation into semantically meaningful components, which cannot be effectively handled by conventional sentence-level segmentation methods. To address this, we propose \task, a novel approach that jointly performs efficient token-level text segmentation and label prediction for long structured texts. Instead of generating full texts for each segment, it generates only starting tokens and reconstructs the complete texts by locating these tokens within the original texts, thereby reducing output tokens by 90\% and minimizing hallucination. To train the models for the boundary generation, \task~performs reinforcement learning with verifiable rewards (RLVR) that jointly optimizes document reconstruction fidelity and semantic alignment. It further mitigates entropy collapse by constructing intermediate candidates by perturbing segment boundaries and labels to create stepping stones toward higher-quality solutions. Experiments show that \task~enables small language models (1.7B parameters) to outperform few-shot prompting with much larger models as well as SFT and standard RLVR baselines on complex prompts used for LLM applications.
\end{abstract}

\section{Introduction}

Text segmentation is the task of dividing a text into coherent segments, each covering a distinct topic \citep{hearst-1994-multi}. Beyond identifying segment boundaries, some approaches also predict the topic of each segment \citep{arnold2019sector}. These segments can help readers better understand the structure of long texts \citep{jeoung2025promptprism}, QA systems retrieve more relevant contexts \citep{ wang2025document}, prompt optimization system more efficiently optimize long prompts \citep{schnabel-neville-2024-symbolic}, and summarization systems summarize long documents \citep{moro2022semantic}.  

 Most previous works \citep{hearst-1994-multi, lukasik-etal-2020-text} perform text segmentation on the sentence or paragraph level. These methods assume that texts can be cleanly divided into sentences or paragraphs. However, this assumption breaks down for many structured texts, such as LLM prompts, that include tables, code snippets, and placeholders. These elements generally do not conform to conventional sentence or paragraph structure.  A natural solution is to adapt these methods to perform text segmentation at the token level. However, sequence labeling on the token level tends to generate very fragmented segments, while boundary classification requires an impractically large number of classifications for each token. Recently, \citet{schnabel2024symbolic} performed token-level segmentation by instructing LLMs to generate the full text of each segment. However, they face high inference costs and hallucination risk for long texts due to regenerating the entire input \citep{wang-etal-2024-positionid}.

\begin{figure*}[t]
\centering
\includegraphics[width=0.98\textwidth, keepaspectratio]{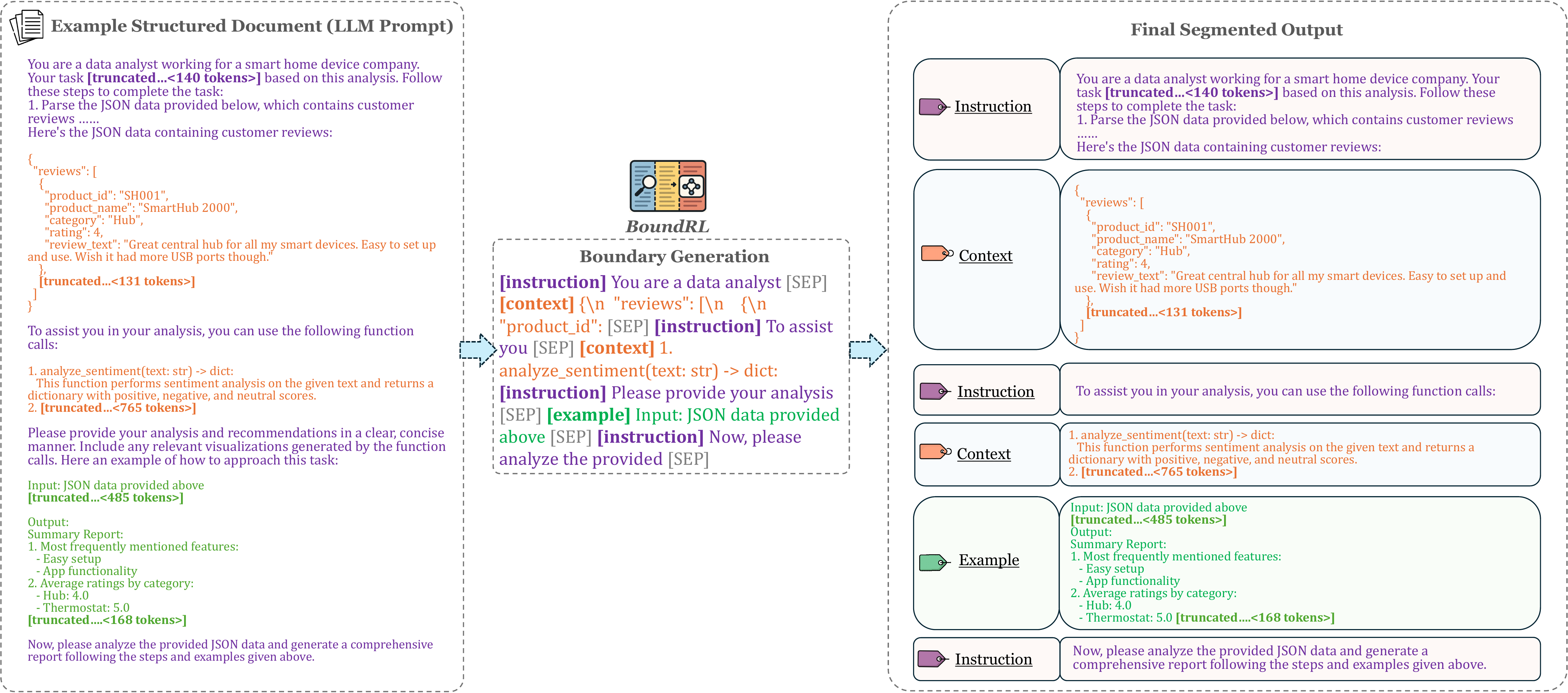}
\caption{Efficient output pattern used by \task. Instead of generating full segment text, it only generates starting tokens for each segment and then reconstructs full segments by locating these tokens in the original text, thereby reducing inference costs by orders of magnitude and minimizing hallucination. Note how the output length is independent of input document length.}
\label{fig:prompt_segmentation}
\end{figure*}

We introduce \task, a novel approach that jointly performs token-level text segmentation and label prediction specifically designed for long structured texts, which we term \emph{structured text segmentation}. As shown in Fig. \ref{fig:prompt_segmentation}, \task~reformulates structured text segmentation as \emph{boundary generation} and only generates a sequence of starting tokens and a label for each segment, then reconstructs full segments by locating them in the input. Unlike methods that generate every segment in full, this formulation reduces output tokens during inference by 90\%, shifting complexity from linear in text length to linear in the number of segments. It also mitigates hallucination risks.

Boundary generation training presents unique optimization challenges. Supervised fine-tuning (SFT) can mistakenly penalize starting tokens that correspond to the right boundary positions and provides insufficient penalties for minor token mismatches that cause failures in locating starting tokens. \task~addresses this by reinforcement learning with verifiable rewards (RLVR) \citep{shao2024deepseekmath}, optimizing a reward function with two complementary dimensions: \emph{reconstruction fidelity}, which measures whether the text can be fully recovered from generated segments, and \emph{semantic alignment} which evaluates agreements between generated segments and annotated segments. This reward design ensures that different starting tokens corresponding to the same boundary receive identical rewards, while starting tokens with minor mismatches receive smaller rewards. 

\begin{figure*}[t]
\centering
\includegraphics[width=0.98\textwidth, keepaspectratio]{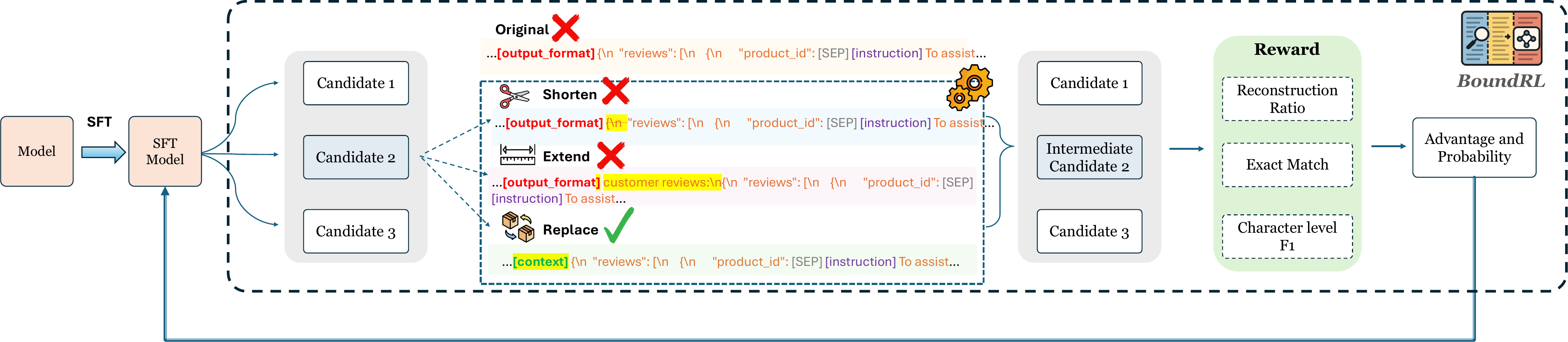}
\caption{RLVR workflow of \task~showing the dual-objective reward function and intermediate candidate construction. The reward function combines reconstruction ratio for reconstruction fidelity with exact match and character-level F1 scores for semantic alignment. To mitigate entropy collapse during rollout, \task~constructs intermediate candidates by perturbing generated segments through boundary adjustments and label modifications.}
\label{fig:BoundRL}
\end{figure*}

However, RLVR can suffer from entropy collapse \citep{cui2025entropy}, where generated sequences of segments become trapped in narrow, low-reward regions during rollout. Although annotated sequences of segments can provide high-reward examples, they are often too distant from the model's current generation distribution to enable effective learning. To mitigate this, \task~constructs intermediate candidates by perturbing generated sequences of segments through boundary adjustments and label modifications as shown in Fig. \ref{fig:BoundRL}, creating stepping stones that bridge the gap between current generations and optimal solutions. This approach is particularly effective for our reward function due to its dense, continuous nature.  

To evaluate \task~on particularly challenging structured texts, we construct \dataset, a comprehensive dataset for structured text segmentation with 15.3K annotations of synthetic prompts and prompts from LangSmith\footnote{https://smith.langchain.com/hub/} with text and label of each segment. Our evaluation focuses on prompts for LLMs due to their extreme structural complexity -- dense mixtures of natural language instructions, code snippets, JSON formatting, and placeholders, making them unsuitable for sentence- or paragraph-level segmentation. 

Our experiments show that relatively small models (1.7B-4B parameters) trained with \task~outperform few-shot prompting using much larger models (Claude-4 Sonnet \citep{claude4}). 
Moreover, \task~brings significant performance and generalization improvement over SFT and the intermediate candidates can further improve the performance of RLVR. 

Our contributions are four-fold:
\begin{itemize}[nosep, leftmargin=10pt]
    \item \task, a boundary-generation approach for token-level structured text segmentation that reduces output tokens by 90\% while mitigating hallucination risks;
    \item A dual-objective reward function optimizing reconstruction fidelity and semantic alignment;
    \item A selective perturbation strategy that addresses entropy collapse by generating learnable intermediate candidates during RLVR;
    \item \dataset, a 15.3K human-annotated benchmark where \task~enables 1.7B models to outperform much larger models such as Claude-4 Sonnet.
\end{itemize}

\section{Related Work}

\noindent\textbf{Text segmentation} Text segmentation aims to divide a text into coherent segments, where each segment encompasses a distinct semantic unit or topic \citep{hearst-1994-multi}. The task has been studied across various domains, such as regular documents \citep{koshorek-etal-2018-text} and dialogues \citep{xing-carenini-2021-improving}. \citet{arnold2019sector,barrow2020joint} extend the task by jointly modeling segmentation and topic classification. In supervised settings, segmentation is often framed as sequence labeling, where each sentence is labeled as a boundary or not \citep{koshorek2018text,li2018segbot}. Other works frame text segmentation as a boundary classification task predicting whether a sentence is a boundary based on its surrounding context \citep{lukasik-etal-2020-text}. More recently, \citet{inan2022structured, duarte2024lumberchunker} frame the text segmentation as a generation task by generating the starting sentence or paragraph indices of each segment. However, these methods face limitations on structured texts containing elements like code snippets and JSON data formats, which lack traditional sentence or paragraph boundaries. \citet{schnabel2024symbolic, jeoung2025promptprism,zhang-etal-2023-seq2seq} propose to segment texts using LLMs by generating the full text of each segment. However, they face high inference costs and hallucination risk for long texts. Given these challenges, \task~frame token-level structured text segmentation as boundary generation. 

\noindent\textbf{Reinforcement Learning with Verifiable Reward} Compared with RLHF, which relies on a separate reward model to assign rewards~\citep{ouyang2022training}, RLVR~\citep{shao2024deepseekmath} uses a rule-based reward function. The design makes RLVR particularly efficient for structured text segmentation, as rewards can be easily computed by comparing generated and annotated segments. However, candidates generated during rollout suffer from being trapped in narrow, low-reward regions, known as entropy collapse~\citep{cui2025entropy}.  To mitigate this, \citet{zhang2025stephint} propose to generate candidates conditioned on the prefix of reference data with varying lengths. However, this requires generating as many candidates as there are segments in the input text, leading to high training costs for structured text segmentation. \citet{yan2025learning, dong2025rl} include reference candidates generated by a reference model, which can be too distant from the model's current generation distribution for effective learning. In contrast, \task~proposes to generate intermediate candidates which are closer to the current distribution of generation.

\section{Problem Statement}

Structured texts refer to texts containing structured elements beyond plain texts, such as code snippets. Structured text segmentation takes a structured text as input and generates a list of segments: 
\vspace{-0.2cm}
\begin{equation}
\begin{aligned}
[(l_1,t_1),...,(l_n,t_n)] &=f(d), \\
s.t.~l_i \neq l_{i-1}, t_i\cap t_{i-1}=\emptyset,&~\forall i=2,...,n
\end{aligned}
\end{equation}
where $f$ denotes the system, $d$ denotes the input structured text, $t_i$ denotes the text of the $i$-th segment and $l_i \in L$ denotes the semantic label of the $i$-th segment. $L$ denotes the set of potential labels for segments, which varies by domains. For simplicity, we use $T^L$ to denote $[(l_i,t_i)^{n}_{i=1}]$.  For our case study on prompts, labels include `instruction', `example', `context', `question', and `output format' as shown in Tab.~\ref{tab:label_set}.

\section{Method}
The training process of \task~consists of two stages: SFT followed by RLVR.  Sec. \ref{sec:output_pattern} describes adaptation of LLMs to an efficient output pattern via SFT. Sec.\ref{sec:reward} describes the reward design of RLVR. Sec. \ref{sec:intermediate} describes the construction of intermediate candidates for the rollout stage of RLVR.

\subsection{Efficient Output Pattern for Structured Text Segmentation}

\label{sec:output_pattern}

To enable efficient structured text segmentation, \task~formulates the task as boundary generation. Instead of regenerating full segment text, it produces only a sequence of starting tokens and a label for each segment, then reconstructs the segments from the input as shown in Fig. \ref{fig:prompt_segmentation}. Specifically, given an input structured text $d$, \task~tunes the LLM to generate a sequence of starting tokens $s_i$ and a label $l_i$ for each segment: $[(\hat{l}_i,\hat{s}_i)_{i=1:n}]$. To adapt LLMs for boundary generation, \task~transforms the annotated text of each segment $t_i$ into corresponding starting token sequences $s_i$. The length of each starting token sequence is randomly sampled between 2 and 10 tokens. To reduce the chance that different segments share the same starting tokens, \task~increases the length of corresponding sequences of starting tokens until they become different or until the full segment length is reached. The LLMs are then fine-tuned using SFT on the starting tokens sequences and their corresponding labels.

\task~then reconstructs the text of each segment $\hat{t}_i$ using the position of each sequence of starting tokens $\hat{s}_i$ in the input structured text $d$. To prevent potential ambiguity among segments that share the same starting tokens, the reconstruction process operates iteratively. Specifically, for the first sequence of starting tokens $\hat{s}_1$, \task~locates it as its leftmost occurrence in the input $d$. For each subsequent sequence $\hat{s}_i$, \task~locates it as its leftmost occurrence in the input $d$ \textbf{after} the position of its previous segment. This introduces an ordering constraint, where each match must occur later in the input text than the previous one. Therefore, even if the same starting token sequence appears multiple times, each occurrence is uniquely assigned to a segment based on its position in this left-to-right traversal. The text of the $i$-th segment $\hat{t}_i$ is then extracted as the text span between sequences of starting tokens $\hat{s}_i$ and $\hat{s}_{i+1}$. For simplicity, we use $\hat{T}^L$ to denote the sequence of reconstructed segment texts and their labels $[(\hat{l}_i,\hat{t}_i)^n_{i=1}]$. If the positions of either $\hat{s}_i$ or $\hat{s}_{i+1}$ cannot be found, the $i$-th segment will be discarded. Compared with regenerating full text of each segment, \task~reduces hallucination risks inherent in text generation while reducing required output tokens from $O(|d|)$ to $O(n)$ tokens, where $|d| \gg n$ denotes the input length.


\subsection{Reward Design of RLVR}
\label{sec:reward}
Boundary generation presents optimization challenges for SFT, as it can mistakenly penalize correct starting tokens and provides a insufficient penalty for minor token mismatches. Therefore, \task~applies RLVR after SFT using the reward function with two dimensions: \textit{reconstruction fidelity}, which measures whether the input can be fully recovered from generated segments, and \textit{semantic alignment} which evaluates agreements between generated and annotated segments. 

To measure reconstruction fidelity, \task~uses the reconstruction ratio $\rho_{\textrm{rec}}$. The metric is calculated as the proportion of the input text $d$ that can be successfully reconstructed from the texts of generated segments $\hat{t}_i$: $\rho_{\textrm{rec}}(\hat{T}^L,d)={\sum^n_{i=1}|\hat{t_i}|}/{|d|}$, where $|*|$ denotes the length of a text in characters. The reconstruction ratio $\rho_{\textrm{rec}}(*)$ ranges from zero to one, with higher values indicating more complete reconstruction of the input structured text $d$. Compared with SFT, the metric imposes stronger penalties on segments with token mismatches. 

To measure semantic alignment, \task~uses two metrics. The first metric is the F-1 score of exact match $\textrm{EM}(\hat{T}^L, T^L)$ between generated segments $\hat{T}^L$ and annotated segments $T^L$ \citep{tjong-kim-sang-de-meulder-2003-introduction}. A generated segment $(\hat{l}_i,\hat{t}_i)$ is considered an exact match to an annotated segment $(l_j,t_j)$ if their texts and labels are the same. The F1 score is then computed as the harmonic mean of precision (the fraction of generated segments an exact match) and recall (the fraction of annotated segments with an exact match). However, the F-1 score of exact match $\textrm{EM}(*)$ can be too strict since minor token-level differences between segment texts can lead to a mismatch. Therefore, \task~additionally uses the character-level F-1 score $\textrm{F1}_{\textrm{char}}(\hat{T}^L, T^L)$ motivated by part-of-speech (POS) tagging \citep{marcus-etal-1993-building}. The metric treats structured text segmentation as a character-level labeling task, where each character in the input $d$ is assigned a label from the set $L$, based on the segment it belongs to. Specifically, all characters in $\hat{t}_i$ are assigned the label $\hat{l}_i$, and likewise for annotated segments. The metric is then calculated as the weighted F-1 score between character-level labels from generated segments and those from annotated segments. Both the F-1 score of exact match $\textrm{EM}(*)$ and the character-level F-1 score $\textrm{F1}_{\textrm{char}}(*)$ range from zero to one, with higher values indicating better alignment between generated and annotated segments. Compared with SFT, different starting tokens that correspond to the same boundary have the same value for both metrics. 

The final reward $r(*)$ for the generated segments $\hat{T}^L$ is calculated using both dimensions:
\vspace{-0.2cm}
\begin{equation}
r(\hat{T}^L)=  \rho_{\textrm{rec}}(\hat{T}^L)\frac{(\textrm{EM}(\hat{T}^L)+  \textrm{F1}_{\textrm{char}}(\hat{T}^L))}{2}
\end{equation}

For simplicity, the equation omits the input text $d$ and the annotated segments $T^L$. The reward encourages generated segments to accurately reproduce starting tokens for complete reconstruction while getting close to annotated segments for high-quality segmentation.

\subsection{Construction of Intermediate Candidate}
\label{sec:intermediate}
In this section, we describe how \task~constructs and incorporates intermediate candidates during the rollout stage of RLVR to mitigate entropy collapse. Specifically, \task~constructs intermediate candidates by perturbing generated candidate segmentations and selectively replaces the originally generated candidate segmentations with the intermediate candidates for the training.

An effective intermediate candidate should lie between generated and annotated segments to provide meaningful yet learnable guidance. To construct such intermediate candidates, \task~first generates $m$ candidate segmentations for an input $d$ using standard RLVR practice, denoted as $[\hat{T}^L_j]_{j=1}^{m}$. These candidate segmentations are ordered by descending reward $r(\hat{T}^L_j)$. \task~then perturbs the candidate segmentation with the medium-level reward: $\hat{T}^L_{\frac{m}{2}}$. As shown in Fig. \ref{fig:BoundRL}, three types of perturbations are considered for each segment: (1) shortening the text $\hat{t}_i$ by truncating one word from either side, (2) extending the text $\hat{t}_i$ by including additional one word from either side, or (3) replacing the label $\hat{l}_i$ with an alternative from the set of potential labels $L$, excluding labels already assigned to neighboring segments. To shorten or extend the text of a segment $\hat{t}_i$, \task~modifies the starting token sequences $\hat{s}_i$ or $\hat{s}_{i+1}$ accordingly. Applying a single perturbation to each segment creates a pool of potential intermediate candidates, each differing from the original by exactly one perturbation. The potential intermediate candidate with the highest reward $r(*)$ is selected as the final intermediate candidate, denoted as $\tilde{T}^{L}_{\frac{m}{2}}$. 

To avoid performance degradation from off-policy intermediate candidates, \task~employs a selective replacement strategy. In each training batch, \task~replaces the original candidate segmentations with the medium-level reward with the intermediate candidate for at most $k$ inputs. Replacement is allowed only when the reward of the intermediate candidate $r(\tilde{T}^{L}_{\frac{m}{2}})$ is higher than the original reward $r(\hat{T}^L_{\frac{m}{2}})$. If more than $k$ inputs satisfy this criterion, \task~selects the top-$k$ with the largest gains to preserve training stability while incorporating the most beneficial refinements. For the selected inputs, \task~uses the intermediate candidate with the remaining $m$-1 generated candidate segmentations in the following training; otherwise, all $m$ generated candidates are used. 

\section{Experimental Setup}
\begin{table}[]
\footnotesize
\centering
\begin{tabular}{@{}lccc@{}}
\toprule
& \textbf{Prompts} & \textbf{Tokens} & \textbf{Segments} \\ 
\midrule
\texttt{Synthetic} & 15,132 & 900 & 6.1 \\
\texttt{Langchain} & 197 & 914 & 7.6 \\
\bottomrule
\end{tabular}
\caption{Statistics of synthetic and real-world prompts.}
\label{tab:data_statistics}
\end{table}
\subsection{\dataset}
In this section, we describe \dataset, which serve as a case study of structured text segmentation. Table~\ref{tab:data_statistics} summarizes the dataset statistics. It contains synthetic and real-world prompts. The synthetic prompts are generated using Claude 3.5 Sonnet \citep{claude3}, balancing diversity and complexity. To ensure diversity, we implement a multi-faceted sampling strategy that draws from varied prompt types (e.g. system prompt), prompt modes (e.g. prompt template), and task types (e.g. classification) and other factors. Additionally, we encourage the generated prompts to include structural elements, such as nested JSON, which make the prompts unsuitable for sentence-level segmentation. Some prompts exceed 2,000 words. More details of synthetic prompts are in App. \ref{app:synthetic_prompt}. The real-world prompts are collected from Langchain-hub \footnote{https://smith.langchain.com/hub} as task templates with each prompt corresponding to a distinct task to ensure diversity.

After collecting synthetic and real-world prompts, we recruit a group of highly experienced human annotators to perform annotation. Each segment is assigned one of five labels: `instruction', `example', `context', `question', or `output format' (Table~\ref{tab:label_set}) following \citet{mao2025prompts, jeoung2025promptprism}. To ensure high annotation quality, we develop step-by-step labeling instructions for annotators. Annotators should first decompose each prompt into mutually exclusive, non-overlapping segments. Placeholders are extracted separately if they are knowledge input, user questions, or contextual information. Then, annotators determine if each segment is an instruction or few-shot examples. If neither applied, annotators write a description of the segment and then select the most appropriate label from the remaining labels.

We denote the subset of synthetic prompts as \texttt{Synthetic} and the subset of real-world prompts as \texttt{Langchain}. Tab. \ref{tab:data_statistics} reports the statistics of these subsets, and Fig. \ref{fig:proportion} shows the distribution of segment labels. For the \texttt{Synthetic} subset, we use 14,732 prompts for training, 200 for validation, and 200 for testing. For the \texttt{Langchain} subset, all prompts are used exclusively for testing.

\subsection{Implementation Details}
We evaluate \task~on three LLMs: Qwen3-1.7b, Qwen3-4b \citep{yang2025qwen3}, and Llama-3.1-8b-Instruct \citep{dubey2024llama}. The training process has two stages: 

\noindent \textbf{Stage 1: SFT} All LLMs are first fine-tuned on the training set of \dataset~for one epoch with a batch size of 16. We use learning rates of 2e-6 for Qwen3 models and 5e-7 for Llama-3.1-8b-Instruct. 

\noindent \textbf{Stage 2: RLVR} SFT-tuned models are then tuned with RLVR using GRPO \citep{shao2024deepseekmath} without standard deviation-based reward scaling \citep{liu2025understanding}. To control computational costs, we use a randomly sampled 25\% subset of the training data. Each training batch contains 6 input documents, with $m=4$ candidate segmentations generated per input text using a temperature of 1.2 during rollout. For intermediate candidate construction, we apply selective replacement with model-specific thresholds: $k=2$ for Qwen3-1.7b and Qwen3-4b, and $k=1$ for Llama-3.1-8b-Instruct. Learning rates are 1e-6 for Qwen3 models and 2e-7 for Llama-3.1. We save checkpoints every 0.2 epochs and select the best model based on validation performance.

During inference, the temperature is set to 0. We tune the hyperparameters based on their performance on the validation set. More implementation details are in App. \ref{app:implementation}. 

\section{Experimental Results}

\subsection{Evaluation of \task}
\label{sec:boundrl}

\begin{table*}[ht!]
\centering
\scriptsize
\setlength{\tabcolsep}{8pt}
\renewcommand{\arraystretch}{1.05}
\begin{tabular}{l|ccccc|ccccc|c}
\toprule
\multirow{2}{*}{\textbf{Method}} & \multicolumn{5}{c}{\textbf{\texttt{Synthetic}}}                                 & \multicolumn{5}{c}{\textbf{\texttt{Langchain}}}                                  & \multicolumn{1}{c}{\multirow{2}{*}{\textbf{Avg}}} \\
                        \cmidrule(lr){2-6} \cmidrule(lr){7-11} 
& $\rho_{\textrm{rec}}$ & $\textrm{EM}$ & $ P_k$  & $\textrm{F1}_{\textrm{lab}}$ & $\textrm{F1}_{\textrm{char}}$ & $\rho_{\textrm{rec}}$ & $\textrm{EM}$ & $ P_k$  & $\textrm{F1}_{\textrm{lab}}$ & $\textrm{F1}_{\textrm{char}}$ & \\ \midrule
Oracle\textsubscript{sent}  & 100.0       & 2.7 & 16.5 & 99.7      & 90.3          & 100.0       & 5.2 & 18.9 & 99.5      & 92.2          & 75.4 \\
ModernBERT+NER & 99.7                                                & 57.3                                                & 5.7                                                & 85.8                                                & 95.2                                                & 98.8                                                & 19.4                                                & 15.6                                                & 66.4                                                & 86.5                                                & 78.8 \\
\rowcolor{gray!15}
\multicolumn{12}{l}{Prompting Baselines} \\
Claude3.5-Sonnet$_{\textrm{full}}$  & 78.3 & 14.3 & 15.4 & 79.8 & 70.2 & 55.8 & 11.0 & 21.4 & 62.4 & 47.5 & 58.2 \\
Claude3.5-Sonnet$_{\textrm{start}}$ & 50.0 & 16.9 & 25.2 & 73.8 & 47.2 & 48.1 & 11.4 & 23.4 & 61.3 & 41.1 & 50.1 \\
Claude4-Sonnet$_{\textrm{full}}$  & 97.2 & 22.1 & 11.3 & 82.2 & 88.2 & 80.3 & 13.8 & 18.1 & 65.5 & 68.3 & 68.8 \\
Claude4-Sonnet$_{\textrm{start}}$ & 90.1 & 22.8 & 13.6 & 79.8 & 81.8 & 87.0 & 18.3 & 18.1 & 67.9 & 71.6 & 68.8 \\

\rowcolor{gray!15}
\multicolumn{12}{l}{Qwen3-1.7b}                                                                                                                                                                     \\
SFT           & 99.3  & 72.3 & 4.6 & 94.1 & 93.7 & 87.7  & 34.7 & 14.7 & 77.0 & 71.1 & 81.1 \\
SFT w/2epochs & 99.5  & 73.5 & 3.9 & 94.4 & 94.6 & 85.0  & 41.0 & 14.6 & 77.6 & 70.9 & 81.8 \\
NER           & 100.0 & 34.1 & 6.5 & 81.5 & 94.8 & 100.0 & 8.9  & 19.9 & 57.4 & 86.7 & 73.7 \\
SFT+RLVR     & 100.0 & 77.4 & 4.1 & 94.7 & 94.6 & 88.6  & 47.2 & 13.2 & 79.1 & 74.6 & 83.9                                     \\
SFT+RLVR$_{\textrm{w/temp.}}$ & 100.0 & 77.2 & 4.0 & 94.6 & 95.0 & 90.4 & 44.6 & 14.4 & 79.1 & 75.4 & 83.8          \\
RL-PLUS                & 99.8  & 73.9 & 4.5 & 94.4 & 94.3 & 91.5 & 42.9 & 13.4 & 79.5 & 76.5 & 83.5          \\
\task                   & 99.9  & 77.3 & 4.1 & 94.8 & 94.8 & 90.6 & 47.3 & 12.2 & 79.8 & 76.8 & \textbf{84.5} \\
\rowcolor{gray!15}
\multicolumn{12}{l}{Qwen3-4b}                                                                                                                       
                                              \\
SFT           & 99.7  & 71.6 & 5.3 & 94.9 & 92.8 & 93.1  & 41.6 & 12.3 & 80.2 & 78.8 & 83.5 \\
SFT w/2epochs & 99.7  & 73.0 & 4.3 & 95.2 & 94.2 & 91.3  & 40.7 & 12.1 & 83.6 & 78.2 & 84.0 \\
NER           & 100.0 & 41.9 & 6.9 & 82.8 & 95.6 & 100.0 & 8.9  & 24.7 & 59.3 & 85.7 & 74.3 \\
SFT+RLVR      & 99.7  & 77.6 & 4.6 & 94.6 & 93.7 & 92.7  & 52.4 & 10.6 & 82.3 & 82.1 & 86.0 \\
SFT+RLVR$_{\textrm{w/temp.}}$ & 99.7 & 77.3 & 4.9 & 94.4 & 93.3 & 87.6 & 47.0 & 12.5 & 77.6 & 74.2 & 83.4          \\
RL-PLUS                & 99.7 & 76.6 & 4.2 & 94.3 & 94.1 & 94.8 & 51.0 & 10.8 & 81.5 & 83.1 & 86.0          \\
\task                   & 99.7 & 78.3 & 4.0 & 94.8 & 94.7 & 94.1 & 52.4 & 10.3  & 82.5 & 83.3 & \textbf{86.6} \\
\rowcolor{gray!15}
\multicolumn{12}{l}{Llama-3.1-8b-Instruct}                                                                                                                                                         \\
SFT           & 99.6  & 71.8 & 4.9  & 94.3 & 93.4 & 95.9 & 28.4 & 13.4 & 79.7 & 80.7 & 82.5 \\
SFT w/2epochs & 99.9  & 72.8 & 4.5  & 94.3 & 94.2 & 95.6 & 31.9 & 13.2 & 78.9 & 79.5 & 82.9 \\
NER           & 100.0  & 25.9 & 12.3 & 69.9 & 92.4 & 100.0 & 7.0  & 24.9 & 55.7 & 83.4 & 69.7 \\
SFT+RLVR      & 100.0 & 73.9 & 4.1  & 94.7 & 94.6 & 96.4 & 40.2 & 11.7 & 77.3 & 82.1 & 84.3     \\ 
SFT+RLVR$_{\textrm{w/temp.}}$ & 99.7  & 72.7 & 4.1 & 94.0 & 94.6 & 91.8 & 43.3 & 13.0 & 77.5 & 78.9 & 83.5          \\
RL-PLUS                & 100.0 & 73.0 & 4.4 & 94.4 & 94.3 & 95.9 & 37.9 & 11.7 & 78.0 & 82.7 & 84.0          \\
\task                   & 100.0 & 76.1 & 4.4 & 94.4 & 94.1 & 96.3 & 42.8 & 11.5 & 78.0 & 82.1 & \textbf{84.8} \\ \bottomrule
\end{tabular}
\caption{Evaluation of \task~across LLMs and datasets. The best-performing method for each LLM is highlighted in \textbf{bold}. \task~consistently outperform both finetuning baselines and few-shot prompting with much larger LLMs. The improvements are particularly big on the \texttt{Langchain} subset, showing \task's superior generalization to real-world, out-of-domain prompts.}
\vspace{-0.45cm}
\label{tab:RLVR}
\end{table*}

In this section, we perform a comprehensive evaluation of \task. We consider the following training schemes: (i) SFT, where models are fine-tuned with SFT for one epoch to adapt the output pattern of \task; (ii) SFT w/2 epochs, where models are fine-tuned with SFT for two epochs; (iii) NER, where models are fine-tuned for two epochs to predict the label for each token like named entity recognition (NER) \citep{li2020survey}; (iv) SFT+RLVR, a two-stage fine-tuning procedure as in \task, but without intermediate candidates; (v) SFT+RLVR$_{\textrm{w/ high temp.}}$, the same as SFT+RLVR but with a higher sampling temperature of $1.5$ during rollout; (vi) RL-PLUS \citep{dong2025rl}, which uses one sequence of annotated segments and three candidate segmentations during rollout. For the NER baseline, we also include a NER baseline fine-tuned on ModernBERT-large \cite{modernbert}, which is the SOTA pretrained model using bidirectional encoder. Additionally, to show the necessity of token-level segmentation instead of sentence-level segmentation for structured text segmentation, we include a sentence-level oracle baseline (\textrm{Oracle\textsubscript{sent}}), where each sentence is mapped to the most overlapping ground-truth label. The \textrm{Oracle\textsubscript{sent}} baseline serves as an upper bound for previous sentence-level segmentation methods \citep{inan2022structured, duarte2024lumberchunker}. Furthermore, we consider a few-shot prompting baseline, where the LLM is instructed to segment an input prompt according to the target taxonomy and a provided example \citep{schnabel-neville-2024-symbolic, zhao-etal-2025-pmpo, jeoung2025promptprism}. For this baseline, we use Claude3.5v2-sonnet \citep{claude3} and Claude4-sonnet \citep{claude4} following \citet{jeoung2025promptprism}.  We consider two output patterns: (i) full, which outputs the complete text of each segment; (ii) start, which outputs only the starting tokens, like \task. Implementation details and qualitative examples of these baselines are in App. \ref{app:baseline} and \ref{app:example} respectively. 

\begin{table*}[]
\centering
\scriptsize
\setlength{\tabcolsep}{8pt}
\renewcommand{\arraystretch}{1.05}
\begin{tabular}{l|ccccc|ccccc|c}
\toprule
\multirow{2}{*}{\textbf{Method}} & \multicolumn{5}{c}{\textbf{\texttt{Synthetic}}}                                 & \multicolumn{5}{c}{\textbf{\texttt{Langchain}}}                                  & \multicolumn{1}{c}{\multirow{2}{*}{\textbf{Avg}}} \\
                        \cmidrule(lr){2-6} \cmidrule(lr){7-11} 
&$\rho_{\textrm{rec}}$ & $\textrm{EM}$ & $ P_k$  & $\textrm{F1}_{\textrm{lab}}$ & $\textrm{F1}_{\textrm{char}}$ & $\rho_{\textrm{rec}}$ & $\textrm{EM}$ & $ P_k$  & $\textrm{F1}_{\textrm{lab}}$ & $\textrm{F1}_{\textrm{char}}$ & \\ \midrule 
\rowcolor{gray!15}
\multicolumn{12}{l}{Qwen3-1.7b}                                                                                                                                                                     \\

\task                   & 99.9  & 77.3 & 4.1 & 94.8 & 94.8 & 90.6 & 47.3 & 12.2 & 79.8 & 76.8 & \textbf{84.5} \\
\task~w/ 2steps          & 99.9  & 78.0 & 4.4 & 94.8 & 94.7 & 88.6 & 45.7 & 13.6 & 80.5 & 75.0 & 83.9          \\
\task~w/o select        & 99.7  & 76.9 & 4.5 & 94.6 & 94.1 & 89.9 & 45.3 & 12.2 & 79.8 & 76.5 & 84.0          \\
\task~w/o middle        & 99.6  & 77.7 & 4.5 & 94.9 & 94.2 & 89.7 & 44.5 & 12.9 & 78.8 & 75.5 & 83.7         \\
\rowcolor{gray!15}
\multicolumn{12}{l}{Qwen3-4b}                                                                                                                                                                       \\
\task                   & 99.7 & 78.3 & 4.0 & 94.8 & 94.7 & 94.1 & 52.4 & 10.3  & 82.5 & 83.3 & \textbf{86.6} \\
\task~w/ 2steps          & 99.7 & 78.3 & 3.7 & 94.5 & 94.7 & 94.7 & 50.0 & 10.9 & 81.5 & 84.8 & 86.4          \\
\task~w/o select        & 99.7 & 77.4 & 4.5 & 94.8 & 93.6 & 93.7 & 52.1 & 10.5 & 83.0 & 84.0 & 86.3          \\
\task~w/o middle        & 99.7 & 77.7 & 4.2 & 94.8 & 94.2 & 94.6 & 50.7 & 11.0 & 81.4 & 84.1 & 86.2                                              \\
\rowcolor{gray!15}
\multicolumn{12}{l}{Llama-3.1-8b-Instruct}                                                                                                                                                         \\

\task                   & 100.0 & 76.1 & 4.4 & 94.4 & 94.1 & 96.3 & 42.8 & 11.5 & 78.0 & 82.1 & \textbf{84.8} \\
\task~w/2steps          & 100.0 & 76.6 & 4.3 & 94.4 & 94.6 & 94.4 & 38.4 & 12.2 & 77.9 & 81.5 & 84.1          \\
\task~w/o select        & 99.9  & 75.7 & 4.3 & 94.3 & 94.6 & 94.7 & 41.8 & 11.6 & 78.7 & 81.5 & 84.5          \\
\task~w/o middle        & 99.9  & 74.7 & 4.4 & 94.0 & 94.2 & 95.5 & 43.2 & 12.1 & 77.6 & 81.8 & 84.4              \\ \bottomrule
\end{tabular}
\caption{Ablation study of \task. The best-performing method of each LLM is in \textbf{bold}.}
\vspace{-0.2cm}
\label{tab:ablation}
\end{table*}

For evaluation, we use the reconstruction ratio $\rho_{\textrm{rec}}(*)$, the F-1 score of exact match $\textrm{EM}(*)$ and the character-level F-1 score $\textrm{F1}_{\textrm{char}}(*)$ as described in Sec. \ref{sec:reward}. We additionally use character-level $P_k$ score \citep{beeferman1999statistical} which measures the quality of segment boundaries, with the window width set to half the average length of annotated segments following standard practice. We also use $\textrm{F1}_{\textrm{lab}}$, which is the micro-F1 score that compares the predicted label of each generated segment with the label of the most overlapping annotated segment following \citet{arnold-etal-2019-sector}. Higher values of all metrics except $P_k$ indicate better performance, while lower $P_k$ values are better. Results are reported in percentage on the test set of the \texttt{Synthetic} subset and the \texttt{Langchain} subset, along with the average of one minus $P_k$ and other metrics across both subsets in Table~\ref{tab:RLVR}. We also evaluate \task~on Wikisection-city dataset \citep{arnold-etal-2019-sector} in App. \ref{app:wikisection}.

We observe that \task~consistently outperforms all baselines. The difference between \task~and the second best-performing method (SFT+RLVR) is statistically significant using paired t-test ($p<0.05$), showing the importance of intermediate candidates for effective RLVR training. We show in App. \ref{app:reward_std} that \task~can mitigate the entropy collapse issue of RLVR in App. \ref{app:reward_std}. In contrast, RL-PLUS, which uses annotated segments during rollout, has inconsistent results and can even hurt performance. This may be because annotated segments are too out-of-distribution to provide useful learning signals. Additionally, increasing temperature (SFT+RLVR$_{\textrm{w/temp.}}$) cannot further improve the performance, showing that the improvement brought by intermediate candidates is not merely from increased exploration space but guided exploration. 

Furthermore, models fine-tuned with RLVR consistently outperform SFT-only models and the improvement becomes bigger on the \texttt{Langchain} subset. The difference in average scores between SFT+RLVR and SFT w/2epochs is statistically significant using paired t-test ($p<0.05$). The results show that SFT is insufficient for optimal performance as it can mistakenly penalize starting tokens that correspond to the right boundaries and give insufficient penalties for minor token mismatches. To further investigate this, we analyze how only generated starting tokens that correspond to the ground-truth positions but differ lexically from the ground-truth tokens for the SFT w/2 epoch baseline. The proportion is 21\% for Llama3.1-8b, 28\% for Qwen3-1.7b, and 19\% for Qwen3-4b. This result indicates that a substantial fraction of predictions identify the correct boundary positions but are penalized due to different tokens.

We note that the smallest model (Qwen3-1.7b) fine-tuned with \task~significantly outperforms the best-performing few-shot prompting baseline (Claude4-sonnet-full) with much more parameters. Besides, prompting baselines that generate full segment text require an average of 1,170 tokens per input prompt on the \texttt{Synthetic} subset, while \task~requires only 119 tokens, which corresponds to a 90\% reduction in output tokens and a significant efficiency gain.

Although models fine-tuned with NER achieve high scores on $\textrm{F1}_{char}$, they generally achieve lower scores on $\textrm{EM}$, $P_k$ and $\textrm{F1}_{lab}$. Analysis of the outputs shows that models fine-tuned with NER tend to generate fragmented and short segments, showing the effectiveness of framing structured text segmentation as a boundary generation task.

Despite having access to gold annotations, Oracle\textsubscript{sent} performs poorly on both \textrm{EM} and $P_k$. This is mainly because LLM prompts contain many code snippets and structured data formats (e.g., JSON) that do not conform to standard sentence boundaries. This demonstrates the necessity of token-level segmentation for structured text segmentation. Moreover, since Oracle\textsubscript{sent} serves as an upper bound for prior sentence-level segmentation methods, its weak performance further shows that these methods are fundamentally inadequate for handling structured text.

\subsection{Ablation Study of \task}

In this section, we perform an ablation study of \task. We consider the following ablated versions of \task: (i) \task~w/ 2steps, which performs two perturbation steps when generating intermediate candidates; (ii) \task~w/o select, which uses an intermediate candidate for all input texts in a batch without selective replacement; (iii) \task~w/o middle, which generates intermediate candidates by perturbing a randomly sampled candidate segmentation instead of the one with the medium-level reward. More details of these ablated versions are in App. \ref{app:ablation}. The results are in Tab. \ref{tab:ablation}.

Tab. \ref{tab:ablation} shows that \task~outperforms all ablated versions. Specifically, applying multiple perturbations when generating intermediate candidates (\task~w/ 2steps) and incorporating them for all input texts (\task~w/o select) both hurt performance. The results show the importance of controlling the distance between the current generation and intermediate candidates, which aligns with our findings in Sec.~\ref{sec:boundrl} that directly using annotated segments does not improve performance.

\subsection{Evaluation of Output Patterns}
\begin{table}[]
\centering
\scriptsize
\begin{tabular}{lccc}
\toprule
                       & start   & end & start+end \\ \midrule
Qwen3-1.7b             & \textbf{81.1} & 75.6      & 74.8            \\
Qwen3-4b               & \textbf{83.5} & 82.3      & 80.3            \\
Llama-3.1-8b-Instruct & \textbf{82.5} & 80.9      & 77.6           \\ \bottomrule
\end{tabular}
\caption{Evaluation of different output patterns. The best-performing output pattern is highlighted in \textbf{bold}. SFT w/start, the output pattern used by \task, consistently outperforms other patterns.}
\label{tab:output_pattern}
\end{table}

In this section, we evaluate output patterns for structured text segmentation by comparing LLMs fine-tuned with SFT for different output patterns. We consider three output patterns: (i) \texttt{start}, which is used by \task~and outputs starting tokens of each segment; (ii) \texttt{end}, which outputs ending tokens of each segment; (iii) \texttt{start+end}, which outputs both starting and ending tokens of each segment. We show examples of these output patterns in App. \ref{app:output_pattern}. We show average scores on Synthetic and Langchain subsets in Tab. \ref{tab:output_pattern} and full results in Tab. \ref{tab:output_pattern_full}.

Tab.~\ref{tab:output_pattern} shows that \texttt{start}, used by \task, consistently outperforms other output patterns across LLMs and datasets, with particular advantages in exact match scores. In contrast, \texttt{start+end} performs worse than both \texttt{start} and \texttt{end}, although it is supposed to be more robust to token mismatches as it generates both boundaries of each segment. This suggests that requiring generation of both starting and ending tokens imposes an excessive learning burden that degrades performance.

\FloatBarrier
\section{Conclusions}
We propose \task, a novel framework that reformulates structured text segmentation as a boundary generation problem. \task~generates only the starting tokens of each segment, which substantially reduces inference costs and mitigates hallucination risks. To train models for boundary generation, \task~employs RLVR to jointly optimize reconstruction fidelity and semantic alignment, while our intermediate candidate construction strategy alleviates entropy collapse during training. In a challenging case study on LLM prompts, \task~consistently outperforms SFT and RLVR baselines, and enables 1.7B models to outperform Claude-4 Sonnet few-shot prompting while using 90\% fewer output tokens. The boundary generation paradigm may extend to other structured text domains such as legal documents and technical specifications. 


\section{Limitations}
While our approach demonstrates promising performance on out-of-distribution structured texts, it still relies on domain-specific annotated datasets and task-specific fine-tuning. This requirement limits its applicability in true zero-shot settings for entirely new document types where annotated data is unavailable. Future work may explore more data-efficient training strategies, such as improved annotation reuse or weak and semi-supervised learning, to further reduce reliance on human-labeled data.


\section{Ethics Statement}
The human annotations were collected through hired annotators from a data annotation service. Annotators were instructed to strictly refrain from including any biased, hateful, or offensive content towards any race, gender, sex, or religion. The annotations underwent audits by a separate group of annotators, achieving an 89\% inter-annotator agreement. We use LLM to polish the writing of the paper.

\bibliography{acl_latex}

@article{ouyang2022training,
  title={Training language models to follow instructions with human feedback},
  author={Ouyang, Long and Wu, Jeffrey and Jiang, Xu and Almeida, Diogo and Wainwright, Carroll and Mishkin, Pamela and Zhang, Chong and Agarwal, Sandhini and Slama, Katarina and Ray, Alex and others},
  journal={Advances in neural information processing systems},
  volume={35},
  pages={27730--27744},
  year={2022}
}

@inproceedings{jeoung2025promptprism,
  title = "{P}rompt{P}rism: A Linguistically-Inspired Taxonomy for Prompts",
    author = "Jeoung, Sullam  and
      Chen, Yueyan  and
      Zhang, Yi  and
      Wang, Shuai  and
      Ding, Haibo  and
      Cheong, Lin Lee",
    editor = "Demberg, Vera  and
      Inui, Kentaro  and
      Marquez, Llu{\'i}s",
    booktitle = "Findings of the {A}ssociation for {C}omputational {L}inguistics: {EACL} 2026",
    month = mar,
    year = "2026",
    address = "Rabat, Morocco",
    publisher = "Association for Computational Linguistics",
    url = "https://aclanthology.org/2026.findings-eacl.61/",
    doi = "10.18653/v1/2026.findings-eacl.61",
    pages = "1168--1192",
    ISBN = "979-8-89176-386-9"
}

@article{mao2025prompts,
  title={From Prompts to Templates: A Systematic Prompt Template Analysis for Real-world LLMapps},
  author={Mao, Yuetian and He, Junjie and Chen, Chunyang},
  journal={arXiv preprint arXiv:2504.02052},
  year={2025}
}

@misc{claude3,
    author    = "Anthropic",
    title     = "Claude 3.5 Sonnet",
    url       = "https://www.anthropic.com/news/claude-3-5-sonnet",
    year      = "2024",
}

@article{shao2024deepseekmath,
  title={Deepseekmath: Pushing the limits of mathematical reasoning in open language models},
  author={Shao, Zhihong and Wang, Peiyi and Zhu, Qihao and Xu, Runxin and Song, Junxiao and Bi, Xiao and Zhang, Haowei and Zhang, Mingchuan and Li, YK and Wu, Yang and others},
  journal={arXiv preprint arXiv:2402.03300},
  year={2024}
}

@inproceedings{tjong-kim-sang-de-meulder-2003-introduction,
    title = "Introduction to the {C}o{NLL}-2003 Shared Task: Language-Independent Named Entity Recognition",
    author = "Tjong Kim Sang, Erik F.  and
      De Meulder, Fien",
    booktitle = "Proceedings of the Seventh Conference on Natural Language Learning at {HLT}-{NAACL} 2003",
    year = "2003",
    url = "https://aclanthology.org/W03-0419/",
    pages = "142--147"
}

@article{marcus-etal-1993-building,
    title = "Building a Large Annotated Corpus of {E}nglish: The {P}enn {T}reebank",
    author = "Marcus, Mitchell P.  and
      Santorini, Beatrice  and
      Marcinkiewicz, Mary Ann",
    editor = "Hirschberg, Julia",
    journal = "Computational Linguistics",
    volume = "19",
    number = "2",
    year = "1993",
    address = "Cambridge, MA",
    publisher = "MIT Press",
    url = "https://aclanthology.org/J93-2004/",
    pages = "313--330"
}

@article{cui2025entropy,
  title={The entropy mechanism of reinforcement learning for reasoning language models},
  author={Cui, Ganqu and Zhang, Yuchen and Chen, Jiacheng and Yuan, Lifan and Wang, Zhi and Zuo, Yuxin and Li, Haozhan and Fan, Yuchen and Chen, Huayu and Chen, Weize and others},
  journal={arXiv preprint arXiv:2505.22617},
  year={2025}
}

@article{dubey2024llama,
  title={The llama 3 herd of models},
  author={Dubey, Abhimanyu and Jauhri, Abhinav and Pandey, Abhinav and Kadian, Abhishek and Al-Dahle, Ahmad and Letman, Aiesha and Mathur, Akhil and Schelten, Alan and Yang, Amy and Fan, Angela and others},
  journal={arXiv e-prints},
  pages={arXiv--2407},
  year={2024}
}

@article{yang2025qwen3,
  title={Qwen3 technical report},
  author={Yang, An and Li, Anfeng and Yang, Baosong and Zhang, Beichen and Hui, Binyuan and Zheng, Bo and Yu, Bowen and Gao, Chang and Huang, Chengen and Lv, Chenxu and others},
  journal={arXiv preprint arXiv:2505.09388},
  year={2025}
}

@misc{claude4,
    author    = "Anthropic",
    title     = "Introducing Claude 4",
    url       = "https://www.anthropic.com/news/claude-4",
    year      = "2025",
}

@article{liu2025understanding,
  title={Understanding r1-zero-like training: A critical perspective},
  author={Liu, Zichen and Chen, Changyu and Li, Wenjun and Qi, Penghui and Pang, Tianyu and Du, Chao and Lee, Wee Sun and Lin, Min},
  journal={arXiv preprint arXiv:2503.20783},
  year={2025}
}

@article{dong2025rl,
  title={RL-PLUS: Countering Capability Boundary Collapse of LLMs in Reinforcement Learning with Hybrid-policy Optimization},
  author={Dong, Yihong and Jiang, Xue and Tao, Yongding and Liu, Huanyu and Zhang, Kechi and Mou, Lili and Cao, Rongyu and Ma, Yingwei and Chen, Jue and Li, Binhua and others},
  journal={arXiv preprint arXiv:2508.00222},
  year={2025}
}

@inproceedings{hearst-1994-multi,
    title = "Multi-Paragraph Segmentation Expository Text",
    author = "Hearst, Marti A.",
    booktitle = "32nd Annual Meeting of the Association for Computational Linguistics",
    month = jun,
    year = "1994",
    address = "Las Cruces, New Mexico, USA",
    publisher = "Association for Computational Linguistics",
    url = "https://aclanthology.org/P94-1002/",
    doi = "10.3115/981732.981734",
    pages = "9--16"
}

@inproceedings{koshorek-etal-2018-text,
    title = "Text Segmentation as a Supervised Learning Task",
    author = "Koshorek, Omri  and
      Cohen, Adir  and
      Mor, Noam  and
      Rotman, Michael  and
      Berant, Jonathan",
    editor = "Walker, Marilyn  and
      Ji, Heng  and
      Stent, Amanda",
    booktitle = "Proceedings of the 2018 Conference of the North {A}merican Chapter of the Association for Computational Linguistics: Human Language Technologies, Volume 2 (Short Papers)",
    month = jun,
    year = "2018",
    address = "New Orleans, Louisiana",
    publisher = "Association for Computational Linguistics",
    url = "https://aclanthology.org/N18-2075/",
    doi = "10.18653/v1/N18-2075",
    pages = "469--473"
}

@inproceedings{xing-carenini-2021-improving,
    title = "Improving Unsupervised Dialogue Topic Segmentation with Utterance-Pair Coherence Scoring",
    author = "Xing, Linzi  and
      Carenini, Giuseppe",
    editor = "Li, Haizhou  and
      Levow, Gina-Anne  and
      Yu, Zhou  and
      Gupta, Chitralekha  and
      Sisman, Berrak  and
      Cai, Siqi  and
      Vandyke, David  and
      Dethlefs, Nina  and
      Wu, Yan  and
      Li, Junyi Jessy",
    booktitle = "Proceedings of the 22nd Annual Meeting of the Special Interest Group on Discourse and Dialogue",
    month = jul,
    year = "2021",
    address = "Singapore and Online",
    publisher = "Association for Computational Linguistics",
    url = "https://aclanthology.org/2021.sigdial-1.18/",
    doi = "10.18653/v1/2021.sigdial-1.18",
    pages = "167--177"
}

@inproceedings{lukasik-etal-2020-text,
    title = "Text Segmentation by Cross Segment Attention",
    author = "Lukasik, Michal  and
      Dadachev, Boris  and
      Papineni, Kishore  and
      Sim{\~o}es, Gon{\c{c}}alo",
    editor = "Webber, Bonnie  and
      Cohn, Trevor  and
      He, Yulan  and
      Liu, Yang",
    booktitle = "Proceedings of the 2020 Conference on Empirical Methods in Natural Language Processing (EMNLP)",
    month = nov,
    year = "2020",
    address = "Online",
    publisher = "Association for Computational Linguistics",
    url = "https://aclanthology.org/2020.emnlp-main.380/",
    doi = "10.18653/v1/2020.emnlp-main.380",
    pages = "4707--4716"
}

@article{arnold2019sector,
  title={SECTOR: A neural model for coherent topic segmentation and classification},
  author={Arnold, Sebastian and Schneider, Rudolf and Cudr{\'e}-Mauroux, Philippe and Gers, Felix A and L{\"o}ser, Alexander},
  journal={Transactions of the Association for Computational Linguistics},
  volume={7},
  pages={169--184},
  year={2019},
  publisher={MIT Press One Rogers Street, Cambridge, MA 02142-1209, USA journals-info~…}
}

@inproceedings{barrow2020joint,
  title={A joint model for document segmentation and segment labeling},
  author={Barrow, Joe and Jain, Rajiv and Morariu, Vlad and Manjunatha, Varun and Oard, Douglas W and Resnik, Philip},
  booktitle={Proceedings of the 58th Annual Meeting of the Association for Computational Linguistics},
  pages={313--322},
  year={2020}
}

@article{inan2022structured,
  title={Structured summarization: Unified text segmentation and segment labeling as a generation task},
  author={Inan, Hakan and Rungta, Rashi and Mehdad, Yashar},
  journal={arXiv preprint arXiv:2209.13759},
  year={2022}
}

@inproceedings{wang-etal-2024-positionid,
    title = "{P}osition{ID}: {LLM}s can Control Lengths, Copy and Paste with Explicit Positional Awareness",
    author = "Wang, Noah  and
      Duan, Feiyu  and
      Zhang, Yibo  and
      Zhou, Wangchunshu  and
      Xu, Ke  and
      Huang, Wenhao  and
      Fu, Jie",
    editor = "Al-Onaizan, Yaser  and
      Bansal, Mohit  and
      Chen, Yun-Nung",
    booktitle = "Findings of the Association for Computational Linguistics: EMNLP 2024",
    month = nov,
    year = "2024",
    address = "Miami, Florida, USA",
    publisher = "Association for Computational Linguistics",
    url = "https://aclanthology.org/2024.findings-emnlp.983/",
    doi = "10.18653/v1/2024.findings-emnlp.983",
    pages = "16877--16915"
}

@article{yan2025learning,
  title={Learning to reason under off-policy guidance},
  author={Yan, Jianhao and Li, Yafu and Hu, Zican and Wang, Zhi and Cui, Ganqu and Qu, Xiaoye and Cheng, Yu and Zhang, Yue},
  journal={arXiv preprint arXiv:2504.14945},
  year={2025}
}

@article{zhang2025stephint,
  title={StepHint: Multi-level Stepwise Hints Enhance Reinforcement Learning to Reason},
  author={Zhang, Kaiyi and Lv, Ang and Li, Jinpeng and Wang, Yongbo and Wang, Feng and Hu, Haoyuan and Yan, Rui},
  journal={arXiv preprint arXiv:2507.02841},
  year={2025}
}

@article{schnabel2024symbolic,
  title={Symbolic prompt program search: A structure-aware approach to efficient compile-time prompt optimization},
  author={Schnabel, Tobias and Neville, Jennifer},
  journal={arXiv preprint arXiv:2404.02319},
  year={2024}
}

@article{beeferman1999statistical,
  title={Statistical models for text segmentation},
  author={Beeferman, Doug and Berger, Adam and Lafferty, John},
  journal={Machine learning},
  volume={34},
  number={1},
  pages={177--210},
  year={1999},
  publisher={Springer}
}

@article{arnold-etal-2019-sector,
    title = "{SECTOR}: A Neural Model for Coherent Topic Segmentation and Classification",
    author = {Arnold, Sebastian  and
      Schneider, Rudolf  and
      Cudr{\'e}-Mauroux, Philippe  and
      Gers, Felix A.  and
      L{\"o}ser, Alexander},
    editor = "Lee, Lillian  and
      Johnson, Mark  and
      Roark, Brian  and
      Nenkova, Ani",
    journal = "Transactions of the Association for Computational Linguistics",
    volume = "7",
    year = "2019",
    address = "Cambridge, MA",
    publisher = "MIT Press",
    url = "https://aclanthology.org/Q19-1011/",
    doi = "10.1162/tacl_a_00261",
    pages = "169--184"
}

@article{li2020survey,
  title={A survey on deep learning for named entity recognition},
  author={Li, Jing and Sun, Aixin and Han, Jianglei and Li, Chenliang},
  journal={IEEE transactions on knowledge and data engineering},
  volume={34},
  number={1},
  pages={50--70},
  year={2020},
  publisher={IEEE}
}

@inproceedings{tiedemann2008simple,
  title={Simple is best: experiments with different document segmentation strategies for passage retrieval},
  author={Tiedemann, J{\"o}rg and Mur, Jori},
  booktitle={Coling 2008: Proceedings of the 2nd workshop on Information Retrieval for Question Answering},
  pages={17--25},
  year={2008}
}

@inproceedings{wang2025document,
  title={Document Segmentation Matters for Retrieval-Augmented Generation},
  author={Wang, Zhitong and Gao, Cheng and Xiao, Chaojun and Huang, Yufei and Si, Shuzheng and Luo, Kangyang and Bai, Yuzhuo and Li, Wenhao and Duan, Tangjian and Lv, Chuancheng and others},
  booktitle={Findings of the Association for Computational Linguistics: ACL 2025},
  pages={8063--8075},
  year={2025}
}

@inproceedings{moro2022semantic,
  title={Semantic self-segmentation for abstractive summarization of long documents in low-resource regimes},
  author={Moro, Gianluca and Ragazzi, Luca},
  booktitle={Proceedings of the AAAI conference on artificial intelligence},
  volume={36},
  number={10},
  pages={11085--11093},
  year={2022}
}

@inproceedings{duarte2024lumberchunker,
  title={LumberChunker: Long-Form Narrative Document Segmentation},
  author={Duarte, Andr{\'e} and Marques, Jo{\~a}o and Gra{\c{c}}a, Miguel and Freire, Miguel and Li, Lei and Oliveira, Arlindo},
  booktitle={Findings of the Association for Computational Linguistics: EMNLP 2024},
  pages={6473--6486},
  year={2024}
}

@inproceedings{koshorek2018text,
  title={Text Segmentation as a Supervised Learning Task},
  author={Koshorek, Omri and Cohen, Adir and Berant, Noam Mor Michael Rotman Jonathan},
  booktitle={Proceedings of NAACL-HLT},
  pages={469--473},
  year={2018}
}

@inproceedings{li2018segbot,
  title={SEGBOT: a generic neural text segmentation model with pointer network},
  author={Li, Jing and Sun, Aixin and Joty, Shafiq},
  booktitle={Proceedings of the 27th International Joint Conference on Artificial Intelligence},
  pages={4166--4172},
  year={2018}
}

@inproceedings{cho2022toward,
  title={Toward Unifying Text Segmentation and Long Document Summarization},
  author={Cho, Sangwoo and Song, Kaiqiang and Wang, Xiaoyang and Liu, Fei and Yu, Dong},
  booktitle={Proceedings of the 2022 Conference on Empirical Methods in Natural Language Processing},
  pages={106--118},
  year={2022}
}

@misc{modernbert,
      title={Smarter, Better, Faster, Longer: A Modern Bidirectional Encoder for Fast, Memory Efficient, and Long Context Finetuning and Inference}, 
      author={Benjamin Warner and Antoine Chaffin and Benjamin Clavié and Orion Weller and Oskar Hallström and Said Taghadouini and Alexis Gallagher and Raja Biswas and Faisal Ladhak and Tom Aarsen and Nathan Cooper and Griffin Adams and Jeremy Howard and Iacopo Poli},
      year={2024},
      eprint={2412.13663},
      archivePrefix={arXiv},
      primaryClass={cs.CL},
      url={https://arxiv.org/abs/2412.13663}, 
}

@inproceedings{zhang-etal-2023-seq2seq,
    title = "Seq2seq is All You Need for Coreference Resolution",
    author = "Zhang, Wenzheng  and
      Wiseman, Sam  and
      Stratos, Karl",
    editor = "Bouamor, Houda  and
      Pino, Juan  and
      Bali, Kalika",
    booktitle = "Proceedings of the 2023 Conference on Empirical Methods in Natural Language Processing",
    month = dec,
    year = "2023",
    address = "Singapore",
    publisher = "Association for Computational Linguistics",
    url = "https://aclanthology.org/2023.emnlp-main.704/",
    doi = "10.18653/v1/2023.emnlp-main.704",
    pages = "11493--11504"
}

@book{bird2009natural,
  title={Natural language processing with Python: analyzing text with the natural language toolkit},
  author={Bird, Steven and Klein, Ewan and Loper, Edward},
  year={2009},
  publisher={" O'Reilly Media, Inc."}
}

@inproceedings{schnabel-neville-2024-symbolic,
    title = "Symbolic Prompt Program Search: A Structure-Aware Approach to Efficient Compile-Time Prompt Optimization",
    author = "Schnabel, Tobias  and
      Neville, Jennifer",
    editor = "Al-Onaizan, Yaser  and
      Bansal, Mohit  and
      Chen, Yun-Nung",
    booktitle = "Findings of the Association for Computational Linguistics: EMNLP 2024",
    month = nov,
    year = "2024",
    address = "Miami, Florida, USA",
    publisher = "Association for Computational Linguistics",
    url = "https://aclanthology.org/2024.findings-emnlp.37/",
    doi = "10.18653/v1/2024.findings-emnlp.37",
    pages = "670--686"
}

@inproceedings{zhao-etal-2025-pmpo,
    title = "{PMPO}: Probabilistic Metric Prompt Optimization for Small and Large Language Models",
    author = "Zhao, ChenZhuo  and
      Liu, Ziqian  and
      Wang, Xinda  and
      Lu, Junting  and
      Ruan, Chaoyi",
    editor = "Christodoulopoulos, Christos  and
      Chakraborty, Tanmoy  and
      Rose, Carolyn  and
      Peng, Violet",
    booktitle = "Findings of the Association for Computational Linguistics: EMNLP 2025",
    month = nov,
    year = "2025",
    address = "Suzhou, China",
    publisher = "Association for Computational Linguistics",
    url = "https://aclanthology.org/2025.findings-emnlp.795/",
    doi = "10.18653/v1/2025.findings-emnlp.795",
    pages = "14728--14761",
    ISBN = "979-8-89176-335-7"
}

@software{vonwerra2020trl,
  title   = {{TRL: Transformers Reinforcement Learning}},
  author  = {von Werra, Leandro and Belkada, Younes and Tunstall, Lewis and Beeching, Edward and Thrush, Tristan and Lambert, Nathan and Huang, Shengyi and Rasul, Kashif and Gallouédec, Quentin},
  license = {Apache-2.0},
  url     = {https://github.com/huggingface/trl},
  year    = {2020}
}

\appendix

\section{Appendix}
\subsection{Taxonomy Used for Prompts}
\begin{table*}[!h]
\small
\centering
\begin{tabular}{ll}
\toprule
Label         & Definition                                                            \\ \midrule
Instruction   & Guidance on how to process and respond to queries.                  \\
Example       & Examples of what the input and corresponding output should look like. \\
Context       & Background information and context that the model needs to refer to.  \\
Question      & Queries or questions provided specifically by users                   \\
Output Format & The type, format, or style of the output      \\ \bottomrule                        
\end{tabular}
\caption{Example taxonomy for structured prompt segmentation used in \dataset. Different domains may employ alternative taxonomies appropriate to their document types and analysis needs.}
\label{tab:label_set}
\end{table*}

In this section, we describe the taxonomy use by \dataset for structured text segmentation. The label of each segment can be `instruction', `example', `context', `question', or `output format' as shown in Tab.~\ref{tab:label_set}

\begin{figure*}[t]
\centering
\includegraphics[width=0.98\textwidth, keepaspectratio]{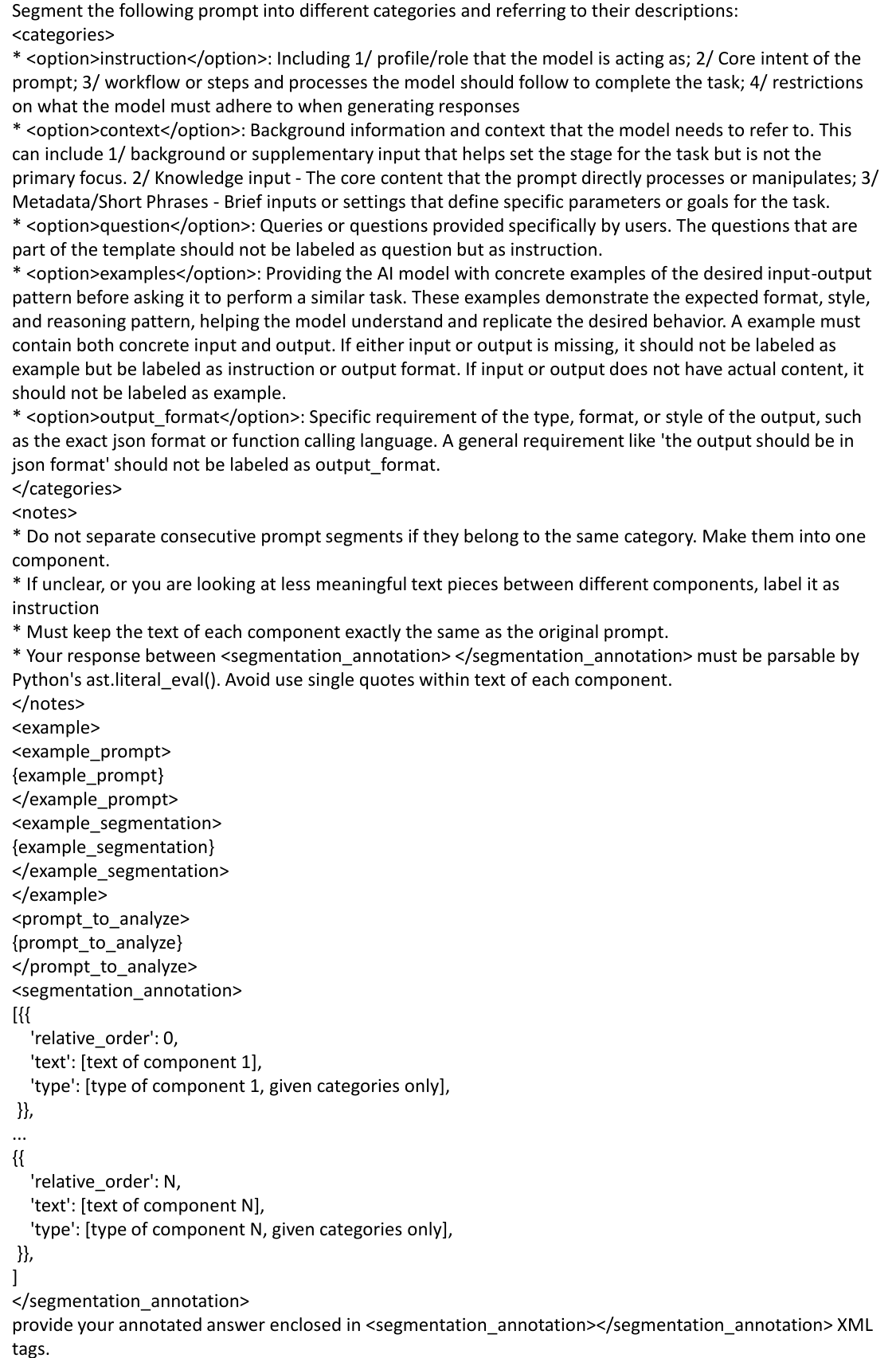}
\caption{Prompt used by Claude to extract the full text of each segment}
\label{fig:prompt_full}
\end{figure*}

\begin{figure*}[t]
\centering
\includegraphics[width=0.98\textwidth, keepaspectratio]{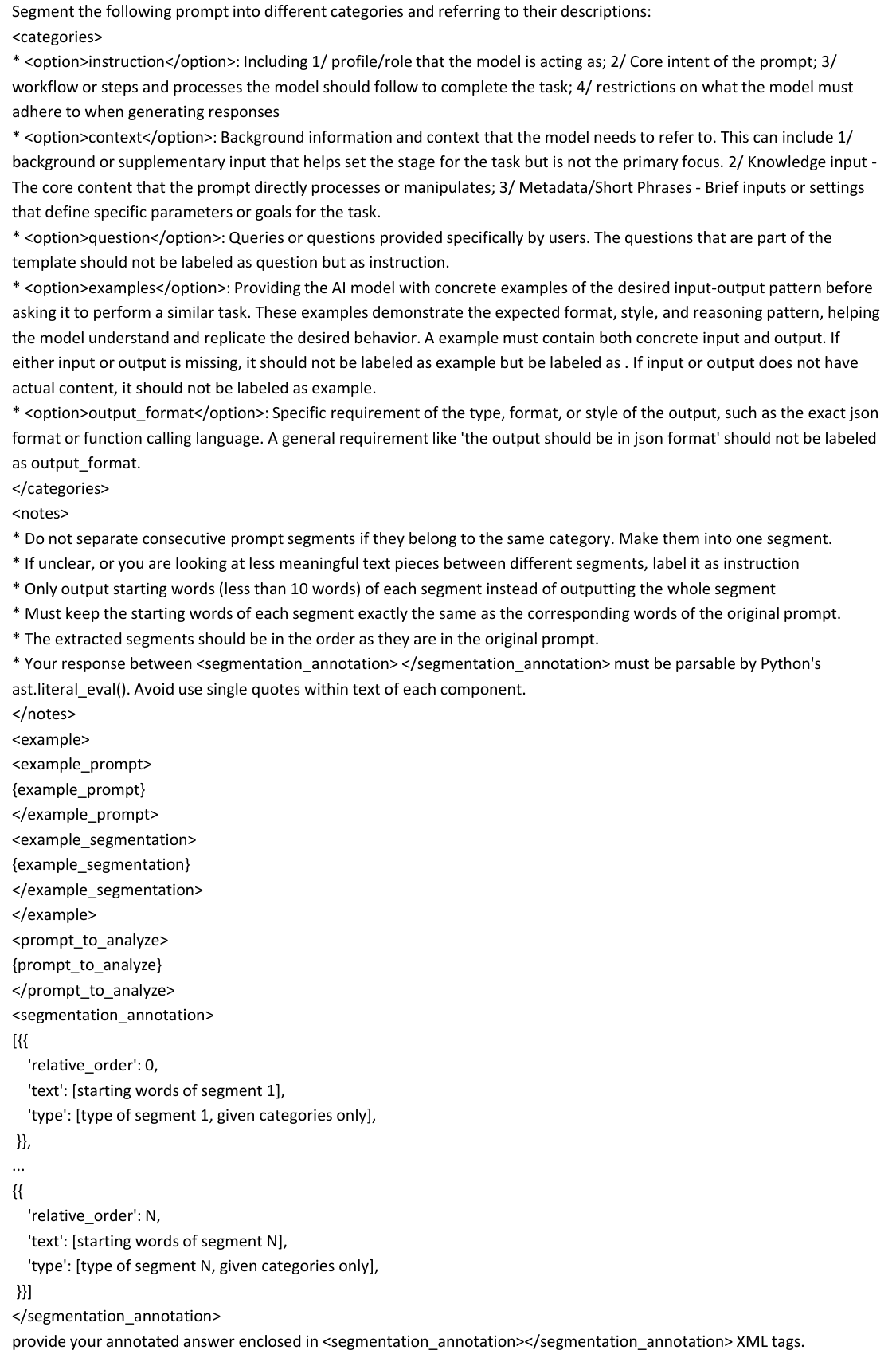}
\caption{Prompt used by Claude to extract the starting tokens of each segment}
\label{fig:prompt_start}
\end{figure*}

\subsection{Implementation Details of \task}
\label{app:implementation}
\begin{figure*}[t]
\centering
\includegraphics[width=0.98\textwidth, keepaspectratio]{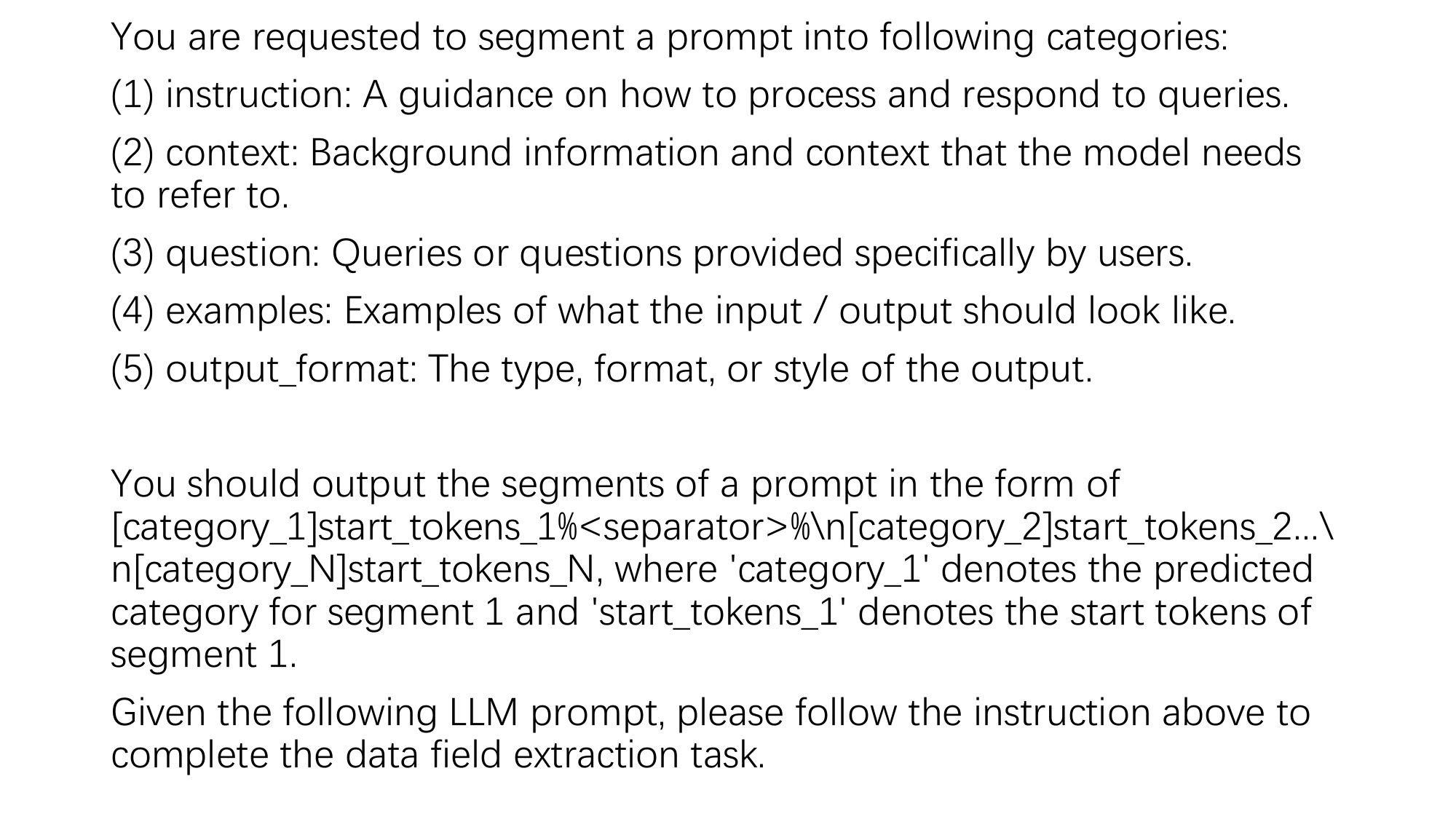}
\caption{Meta instruction used by \task~to output starting tokens of each segment.}
\label{fig:boundrl_start}
\end{figure*}

In this section, we describe additional implementation details of \task. To help model better adapt to the prompt segmentation task, we give models a meta instruction in addition to the prompt to be segmented. The meta instruction includes a brief definition of each segment type and an example of required output format. The meta instruction that requires the model to output starting tokens of each segment is shown in Fig. \ref{fig:boundrl_start}. 

\begin{figure*}[t]
\centering
\includegraphics[width=0.98\textwidth, keepaspectratio]{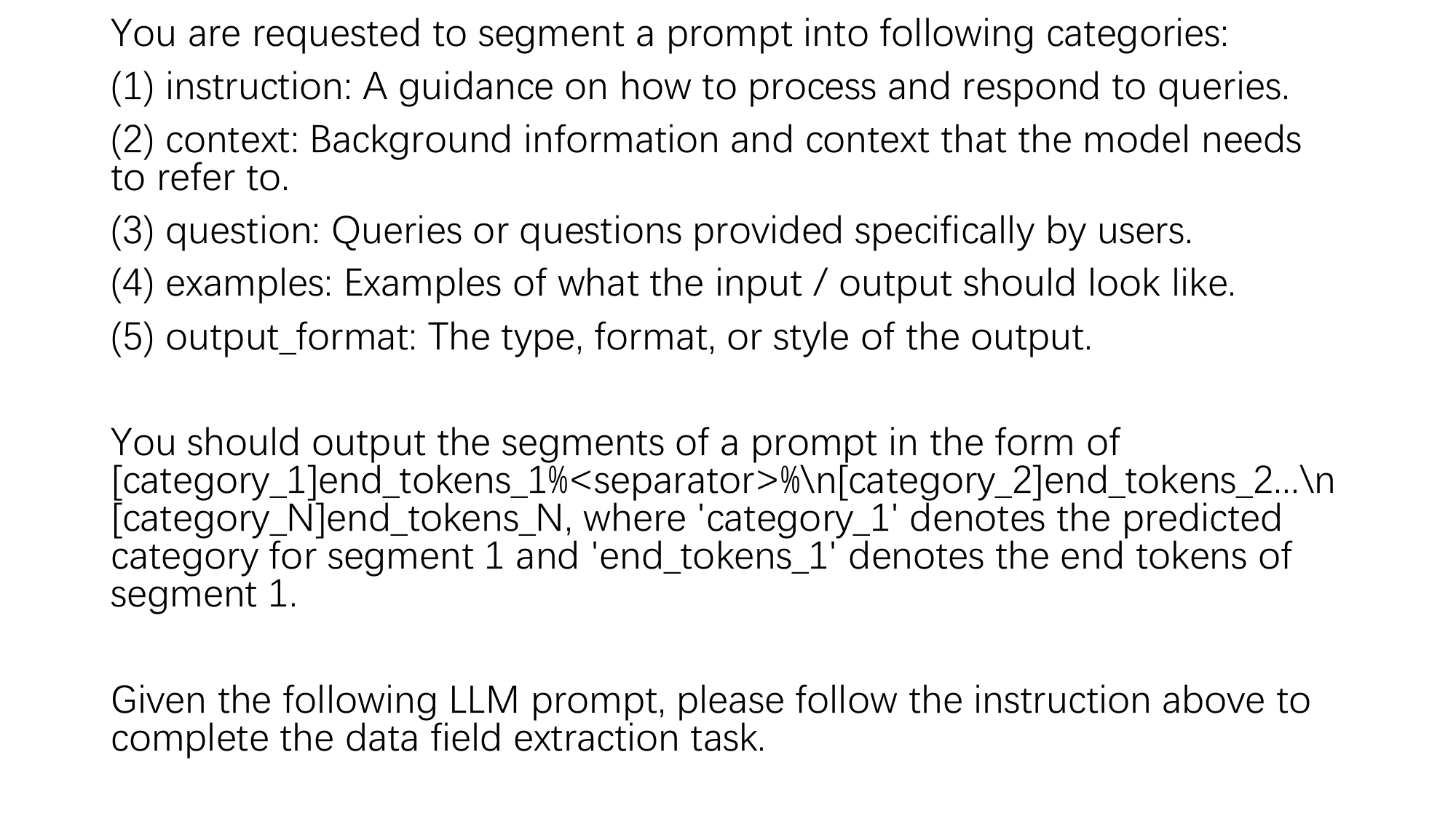}
\caption{Meta instruction used by \task~to output ending tokens of each segment.}
\label{fig:boundrl_end}
\end{figure*}

\begin{figure*}[t]
\centering
\includegraphics[width=0.98\textwidth, keepaspectratio]{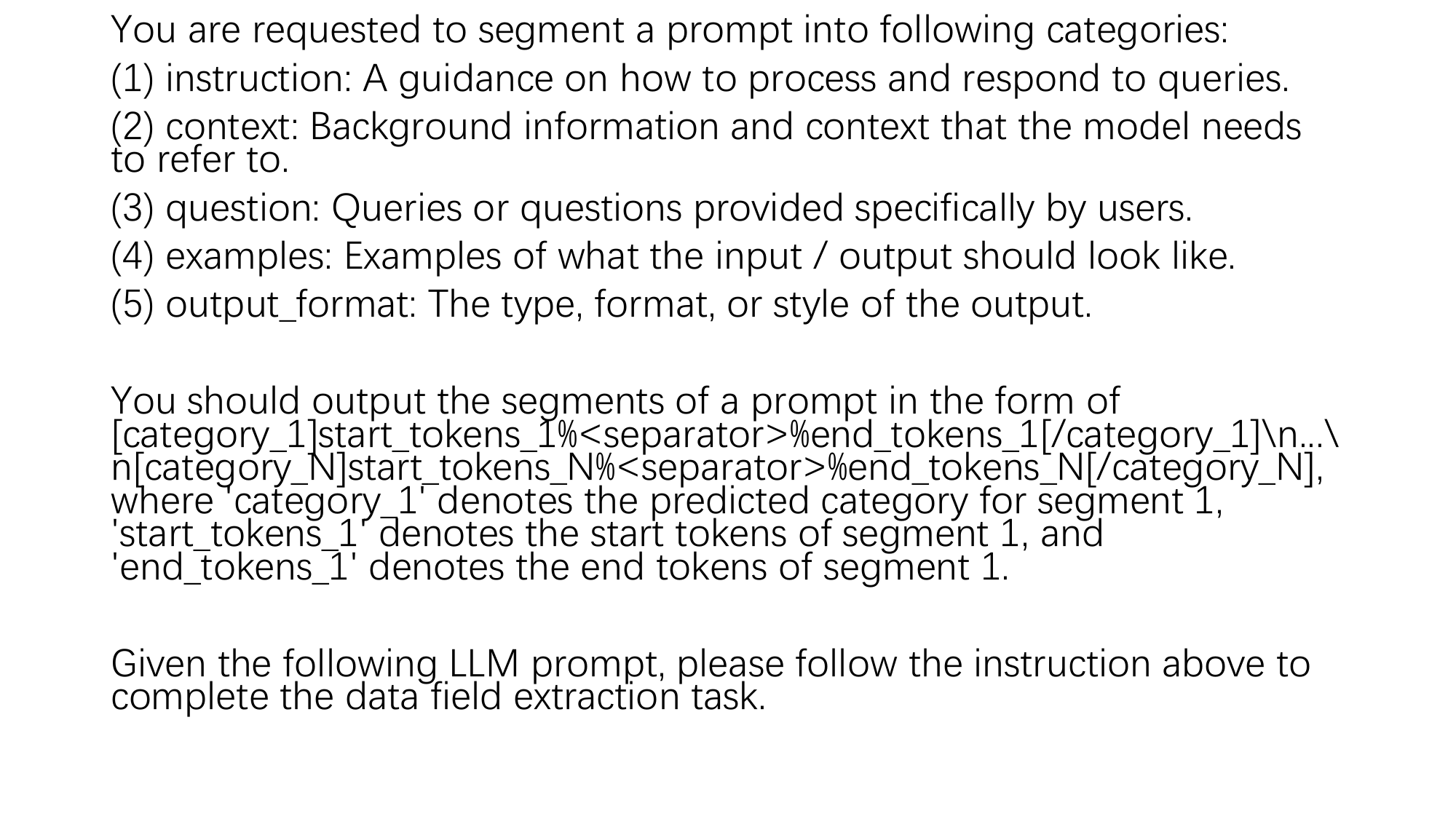}
\caption{Meta instruction used by \task~to output both starting and ending tokens of each segment.}
\label{fig:boundrl_startend}
\end{figure*}

For both SFT and reinforcement learning, \task~sets the maximum gradient norm as 0.1 and the weight decay as 0.01. \task~uses the linear learning rate scheduler with a warmup ratio of 0.03. We tune the hyperparameters of \task~in stages. We first select the hyperparameters of SFT based on the performance of models that are fine-tuned SFT on the validation set. With the SFT hyperparameters fixed, we then tune the hyperparameters of RLVR, followed by those for intermediate candidate construction, using the same process. The SFT and reinforcement learning are implemented using `SFTTrainer' and `GRPOTrainer' from the `trl' package \cite{vonwerra2020trl}. 

During the rollout stage of reinforcement learning, we notice that candidate segmentation might contain repetitive end of response tokens or segments after end of response tokens, which might hurt the training stability if are directly used for training. To address the issue, all generated candidate segmentations are truncated at the first end of response token. 

To shorten or extend the text of a segment $\hat{t}_i$, \task~modifies the starting token sequences $\hat{s}_i$ or $\hat{s}_{i+1}$ accordingly. Specifically, to shorten the text of a segment $\hat{t}_i$ on the left side by one word, \task~truncates the first word of the starting token sequence $\hat{s}_i$. To shorten the text of a segment $\hat{t}_i$ on the right side by one word, \task~prepends to $\hat{s}{i+1}$ the word immediately before it. To extend the text of a segment $\hat{t}_i$ on the right side by one word, \task~truncates the first word of the starting token sequence $\hat{s}_{i+1}$. To extend the text of a segment $\hat{t}_i$ on the left side by one word, \task~prepends to $\hat{s}{i}$ the word immediately before it. Therefore, When constructing intermediate candidates, \task does not shorten or extending the text of a segment with only one word. We will also modify the neighboring starting token sequences accordingly if there is a overlap between starting token sequences after modifications.

\begin{figure*}[t]
\centering
\includegraphics[width=0.9\textwidth, keepaspectratio]{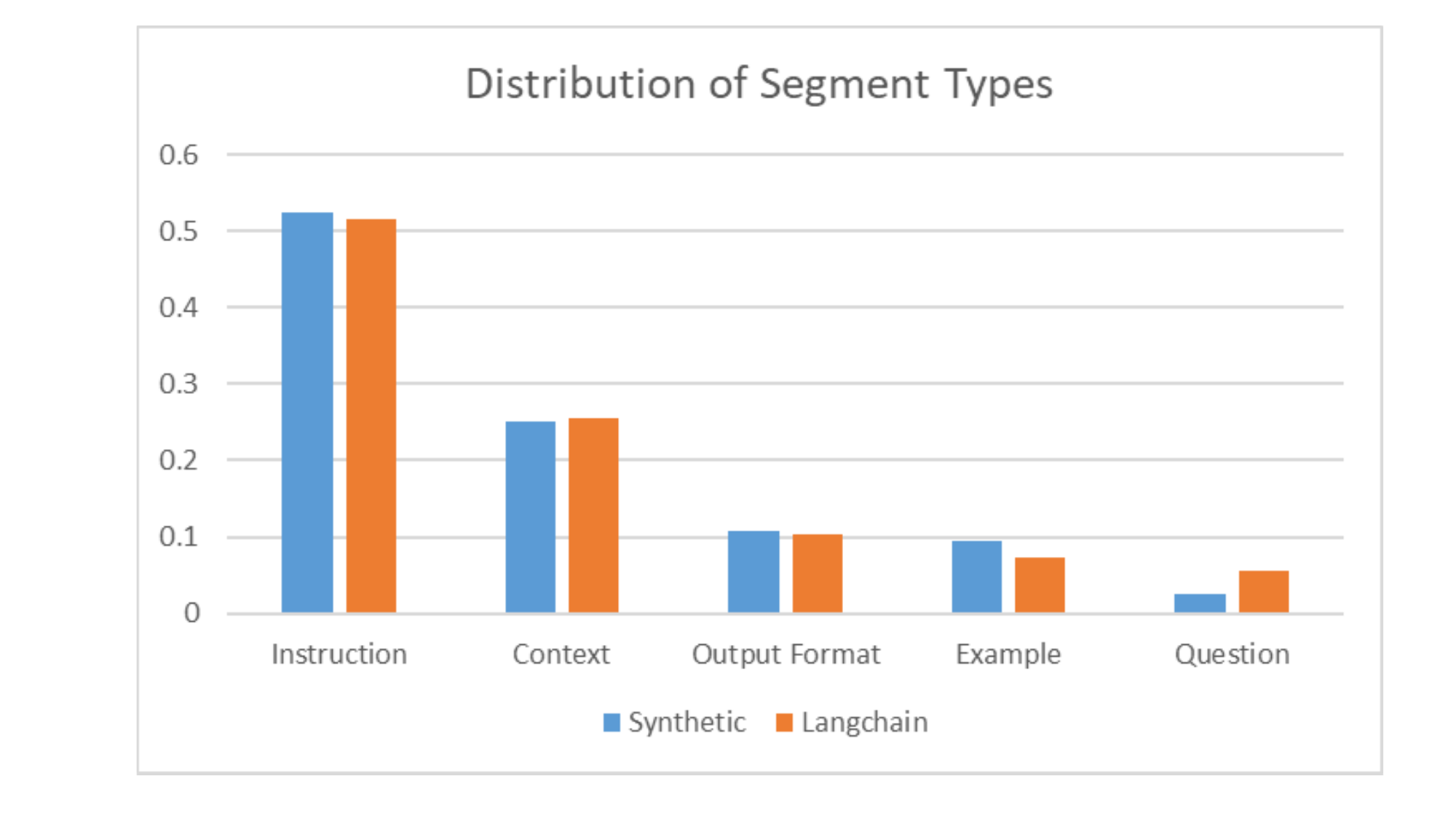}
\caption{Distribution of segment labels across our dataset showing the proportion of each label type in both synthetic prompts (\texttt{Synthetic}) and real-world prompts (\texttt{Langchain}). }
\label{fig:proportion}
\end{figure*}

\subsection{Implementation Details of Baselines}
\label{app:baseline}
In this section, we describe the implementation details of all baselines. For the prompting baseline using Claude3.5-sonnet-v2 and Claude4-sonnet, the prompts used by the models first give a detailed definition for each label used by our taxonomy and then instructs the model to extract segments following the definitions. As described in Sec. \ref{sec:boundrl}, the models are required to output either the full text of each segment or the starting tokens of each segment as \task. To help model better understand the required output format, the prompts also include a randomly sampled prompt and its expected output format from the training set of \dataset. The prompt for outputting the full text of each segment is shown in Figs \ref{fig:prompt_start}. The prompt for outputting the starting tokens of each segment is shown in Fig. \ref{fig:prompt_start}. The temperature during inference is set to 0. 

SFT and SFT w/2epochs use the same hyper-parameters as the SFT training stage of \task. The label schema of our NER baseline is motivated by \citep{tjong-kim-sang-de-meulder-2003-introduction}, which uses `B-X', `I-X', and `O'. However, since all tokens in a prompt belong to a segment in the text segmentation task, we use `B-X' to represent the beginning token of each segment and `I-X' to represent the remaining tokens of each segment. To predict the label of each token, the output of the final layer for each token is feed into a single-layer MLP following the common practice. Other hyper-parameters of the NER baseline is the same as the SFT training stage of \task. For ModernBERT-Large+NER, it is tuned for two epochs with a learning rate of 1e-5. For RL-PLUS, we replace one originally generated candidate segmentation with a sequence of annotated segments. The temperature to control the weight of advantage function is $1.0$. Unless otherwise specified, RL-PLUS, SFT+RLVR, and SFT+RLVR$_{\textrm{w/ high temp.}}$ all use GRPO without reward scaling based on standard deviation and the same hyper-parameters as \task~for a fair comparison.

For \textrm{Oracle\textsubscript{sent}}, we first split the input text $d$ into sentences using `sent\_tokenize' function from the nltk package \cite{bird2009natural}. We then label each sentence as the ground-truth label with the maximum number of overlapping tokens. The final segment is obtained by merging neighboring sentences with the same label.

\subsection{Implementation Details of Other Output Patterns}
\label{app:output_pattern}
\begin{figure*}[t]
\centering
\includegraphics[width=0.9\textwidth, keepaspectratio]{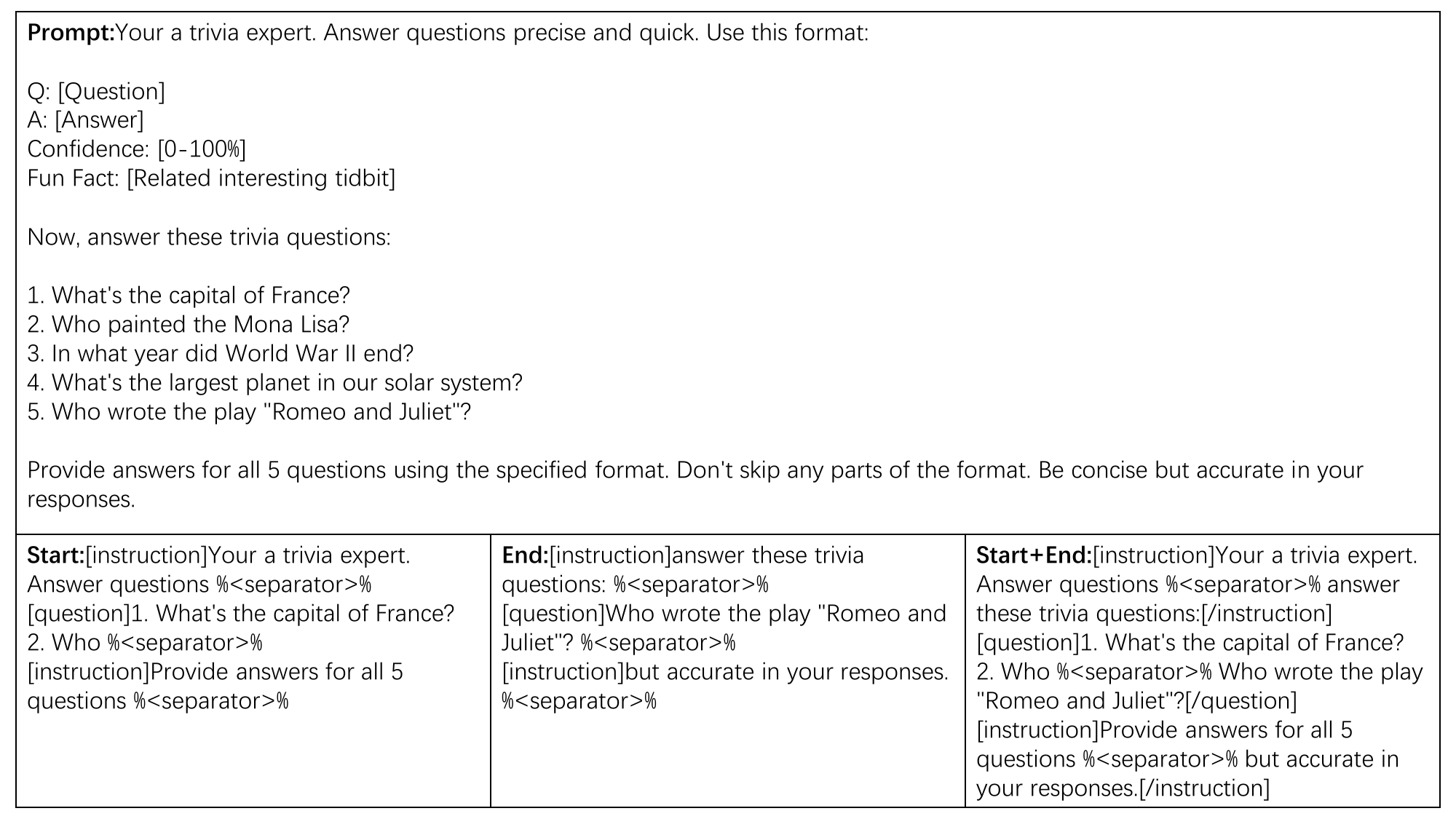}
\caption{An example text and expected outputs for different output patterns. }
\label{fig:output_pattern}
\end{figure*}

In this section, we provide implementation details for output patterns other than the output pattern used by \task~(`start'). Specifically, for the `end' output pattern, the model should first generate a label and then a sequence of ending tokens for each segment.  The text of the $i$-th segment $\hat{t}_i$ is then extracted as the text span between the positions of the $i-1$-th and $i$-th sequence of ending tokens.
For the `start+end', the model should first output a label and then output a sequence of starting tokens and a sequence of ending tokens.  The text of the $i$-th segment $\hat{t}_i$ is then extracted as the text span between the positions of the $i$-th sequence of starting tokens and the $i$-th sequence of ending tokens. We show the meta instruction that requires the model to output ending tokens of each segment in Fig. \ref{fig:boundrl_end} and meta instruction that requires the model to output both starting and ending tokens of each segment in Fig. \ref{fig:boundrl_startend}. We also show an example text and corresponding expected outputs for different output patterns in Fig. \ref{fig:output_pattern}. Furthermore, we also show the full results for different output patterns in Tab. \ref{tab:output_pattern_full}. 

\begin{table*}[t]
\centering
\scriptsize
\setlength{\tabcolsep}{8pt}
\renewcommand{\arraystretch}{1.05}
\begin{tabular}{l|ccccc|ccccc|c}
\toprule
\multirow{2}{*}{\textbf{Method}} & \multicolumn{5}{c}{\textbf{\texttt{Synthetic}}}                                 & \multicolumn{5}{c}{\textbf{\texttt{Langchain}}}                                  & \multicolumn{1}{c}{\multirow{2}{*}{Avg}} \\
                        \cmidrule(lr){2-6} \cmidrule(lr){7-11} 
& $\rho_{\textrm{rec}}$ & $\textrm{EM}$ & $ P_k$  & $\textrm{F1}_{\textrm{lab}}$ & $\textrm{F1}_{\textrm{char}}$ & $\rho_{\textrm{rec}}$ & $\textrm{EM}$ & $ P_k$  & $\textrm{F1}_{\textrm{lab}}$ & $\textrm{F1}_{\textrm{char}}$ & \\ \midrule
\rowcolor{gray!15}
\multicolumn{12}{l}{Qwen3-1.7b}                                                                                                                                                                     \\
SFT w/start             & 99.3                 & 72.3 & 4.6 & 94.1      & 93.7          & 87.7                 & 34.7 & 14.7 & 77.0      & 71.1          & \textbf{81.1}                            \\
SFT w/end               & 96.0                 & 64.8 & 6.0 & 92.0      & 89.3          & 77.7                 & 26.5 & 19.5 & 71.1      & 63.9          & 75.6                                     \\
SFT w/start+end         & 96.6                 & 59.5 & 7.9 & 90.8      & 86.7          & 84.5                 & 20.2 & 17.9 & 69.9      & 65.2          & 74.8                                     \\
\rowcolor{gray!15}
\multicolumn{12}{l}{Qwen3-4b}                                                                                                                                                                       \\
SFT w/start             & 99.7                 & 71.6 & 5.3 & 94.9      & 92.8          & 93.1                 & 41.6 & 12.3 & 80.2      & 78.8          & \textbf{83.5}                            \\
SFT w/end               & 98.1                 & 67.8 & 4.6 & 93.7      & 92.5          & 91.5                 & 39.9 & 13.0 & 79.0      & 78.5          & 82.3                                     \\
SFT w/start+end         & 98.8                 & 70.4 & 5.9 & 93.9      & 92.7          & 85.2                 & 33.7 & 14.1 & 76.5      & 71.8          & 80.3                                     \\
\rowcolor{gray!15}
\multicolumn{12}{l}{Llama-3.1-8b-Instruct}                                                                                                                                                         \\
SFT w/start             & 99.6                 & 71.8 & 4.9 & 94.3      & 93.4          & 95.9                 & 28.4 & 13.4 & 79.7      & 80.7          & \textbf{82.5}                            \\
SFT w/end               & 98.7                 & 65.9 & 5.6 & 93.8      & 92.5          & 93.5                 & 31.5 & 13.6 & 75.6      & 76.6          & 80.9                                     \\
SFT w/start+end         & 97.8                 & 62.9 & 6.2 & 90.8      & 91.4          & 87.7                 & 25.1 & 17.1 & 71.4      & 72.3          & 77.6     \\ \bottomrule
\end{tabular}
\caption{Evaluation of different output patterns. The best-performing output pattern is highlighted in \textbf{bold}. SFT w/start, the output pattern used by \task, consistently outperforms other patterns.}
\label{tab:output_pattern_full}
\end{table*}
\subsection{Generation of Synthetic Prompts}
\label{app:synthetic_prompt}
When generating the synthetic prompts, we implement a multi-faceted sampling strategy that draws from varied prompt types (system prompt, user prompt, combined), prompt modes (prompt template, full prompt, hybrid), and task types (e.g. classification, summarization), placeholder formats (e.g., \texttt{\{context\}} or \texttt{\{\{context\}\}}), the number of examples (zero, one, few ), prompt lengths, writing styles, and levels of details. The full list of the task types, writing styles, prompt lengths, format types, and level of details are shown in Tab. \ref{tab:list_synthetic}.
\begin{table*}[h]
\scriptsize
\begin{tabularx}{\textwidth}{|l|X|}
\hline
Factor          & Values \\ \hline
task type       & 'named entity recognition', 'current events knowledge', 'document similarity comparison', 'bug detection', 'grammar and spell checking', 'code refactoring', 'parameter extraction', 'software development', 'webhook handling', 'clinical note summarization', 'table relationship inference', 'trivia answering', 'anything you can think that a user might need help with', 'web development', 'general productivity assistant', 'spam detection', 'scientific concept explanation', 'mathematical problem solving', 'email thread summarization', 'multi-option reasoning', 'genre classification', 'research paper abstracting', 'classification', 'database schema understanding', 'general coding assistance', 'text simplification', 'keyword extraction', 'news categorization', 'multi-API orchestration', 'fact-checking', 'sentiment analysis', 'citation finding', 'meeting minutes generation', 'news article summarization', 'fact verification', 'legal document summarization', 'complex query generation', 'toxicity detection', 'factoid QA', 'technical domain QA', 'semantic search', 'geographic knowledge QA', 'logical reasoning tasks', 'text generation', 'text style transfer', 'function calling', 'contextual recommendations', 'language detection', 'api authentication', 'evidence extraction', 'table-based QA', 'data filtering', 'code review', 'biographical information retrieval', 'open\_book\_qa (RAG), where a document is provided and a quesiton must be answered, but don\textbackslash{}'t explicitly mention "open book qa"', 'dialogue summarization', 'topic classification', 'code explanation', 'code security enhancements', 'data type validation', 'anomaly detection in text', 'common sense reasoning', 'code generation', 'metaphor generation and interpretation', 'document type classification', 'ai coding assistant', 'poetry and song lyrics generation', 'customer feedback summarization', 'general programming ai assistance', 'code completion', 'chart/graph interpretation', 'automated essay scoring', 'paraphrasing', 'closed\_book\_qa, but don\textbackslash{}'t explicitly mention "closed book qa"', 'multi-hop reasoning', 'input sanitzation', 'content moderation', 'SQL query optimization', 'api endpoint selection', 'context-dependent reasoning', 'error handling', 'intent classification', 'summarization', 'emotion classification', 'cross-document QA', 'multi-document summarization', 'text2sql', 'text completion', 'code summarization', 'reading comprehension', 'historical fact retrieval', 'video transcript summarization'   \\ \hline
writing styles  & 'to contain several noticeable grammatical errorsin direct and curt way', 'to contain several noticeable grammatical errors', 'in direct and curt way', 'to have lots of typos', 'in a well-formed style' \\ \hline
format type     & 'a mixture of markdown and formatting seen in the example prompts provided above', 'a mixture of a formatting structure of your choice and section subtitles', 'a mixture of JSON or nested JSON and other', 'markdown', 'a mixture of JSON or nested JSON and markdown', 'YAML-style formatting', 'a mixture of a formatting structure of your choice and other', 'a mixture of markdown and section headers', 'a mixture of formatting seen in the example prompts provided above and section subtitles', 'a mixture of other and markdown', 'a mixture of XML tags and coding', 'a mixture of coding and XML tags', 'a mixture of section subtitles and a formatting structure of your choice', 'a mixture of JSON or nested JSON and coding', 'a mixture of a formatting structure of your choice and markdown', 'a mixture of section subtitles and coding', 'a mixture of formatting seen in the example prompts provided above and JSON or nested JSON', 'a mixture of XML tags and section subtitles', 'a mixture of XML tags and formatting seen in the example prompts provided above', 'a mixture of section headers and formatting seen in the example prompts provided above', 'a mixture of a formatting structure of your choice and coding', 'pseudo-code', 'a mixture of coding and a formatting structure of your choice', 'a mixture of coding and JSON or nested JSON', 'a mixture of markdown and coding', 'a mixture of other and JSON or nested JSON', 'a mixture of XML tags and a formatting structure of your choice', 'section subtitles', 'a mixture of other and a formatting structure of your choice', 'a mixture of section headers and section subtitles', 'a mixture of other and coding', 'a mixture of section subtitles and markdown', 'a mixture of section subtitles and other', 'a mixture of coding and section headers', 'a mixture of section headers and a formatting structure of your choice', 'a mixture of section headers and coding', 'chain-of-thought styling', 'capital letters to highlight important details', 'a mixture of other and section subtitles', 'a mixture of XML tags and JSON or nested JSON', 'XML tags', 'a mixture of coding and other', 'a mixture of other and formatting seen in the example prompts provided above', 'a mixture of formatting seen in the example prompts provided above and section headers', 'a mixture of a formatting structure of your choice and XML tags', 'a mixture of formatting seen in the example prompts provided above and other', 'a formatting structure of your choice', 'a mixture of section headers and XML tags', 'a mixture of XML tags and markdown', 'a mixture of JSON or nested JSON and a formatting structure of your choice', 'a mixture of markdown and a formatting structure of your choice', 'JSON or nested JSON', 'a mixture of a formatting structure of your choice and section headers', 'a mixture of JSON or nested JSON and section headers', 'a mixture of coding and formatting seen in the example prompts provided above', 'a mixture of JSON or nested JSON and formatting seen in the example prompts provided above', 'a mixture of section subtitles and section headers', 'a mixture of JSON or nested JSON and XML tags', 'a mixture of other and XML tags', 'a mixture of XML tags and other', 'a mixture of section headers and JSON or nested JSON', 'a mixture of markdown and JSON or nested JSON', 'a mixture of a formatting structure of your choice and formatting seen in the example prompts provided above', 'a mixture of section subtitles and XML tags', 'table-based formatting', 'a mixture of a formatting structure of your choice and JSON or nested JSON', 'a mixture of section headers and other', 'a mixture of formatting seen in the example prompts provided above and markdown', 'a mixture of formatting seen in the example prompts provided above and a formatting structure of your choice', 'a mixture of section headers and markdown', 'a mixture of markdown and section subtitles', 'a mixture of markdown and XML tags', 'tree-style hierarchical formatting', 'a mixture of section subtitles and formatting seen in the example prompts provided above', 'a mixture of coding and markdown', 'a mixture of section subtitles and JSON or nested JSON', 'coding', 'a mixture of other and section headers', 'a mixture of coding and section subtitles', 'a mixture of markdown and other', 'formatting seen in the example prompts provided above', 'a mixture of XML tags and section headers', 'a mixture of formatting seen in the example prompts provided above and XML tags', 'section headers', 'a mixture of JSON or nested JSON and section subtitles', 'a mixture of formatting seen in the example prompts provided above and coding' \\ \hline
prompt length   & 'less than 150 words', '150 to 500 words', '500 to 1000 words',  'around 1000 words',  '1000 to 2000 words'  \\ \hline
level of detail & 'basic level of detail, meaning it can just give minimal descriptions of things', 'moderate level of detail, meaning it goes a bit in-depth into things', 'detailed, meaning you should describe things thoroughly and do not give short names or descriptions', 'very detailed, meaning everything is described very in-depth and production-level detail is included', 'extremely technically detailed, meaning as many specific details should be present as possible, including technical jargon, production-level of context, and complicated descriptions'    \\  \hline   
\end{tabularx}
\caption{Full list of factors and corresponding potential values used for generating of synthetic prompts.}
\label{tab:list_synthetic}
\end{table*}
\subsection{Standard Deviation of Rewards during Training}
\label{app:reward_std}
\begin{figure*}[h!]
    \centering
    \begin{subfigure}[]{0.3\textwidth}
        \centering
        \includegraphics[width=\textwidth]{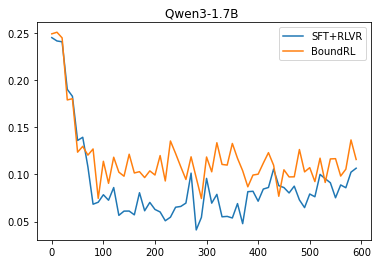}
        \caption{Qwen3-1.7b}
    \end{subfigure}
    \hfill
    \begin{subfigure}[]{0.3\textwidth}
        \centering
        \includegraphics[width=\textwidth]{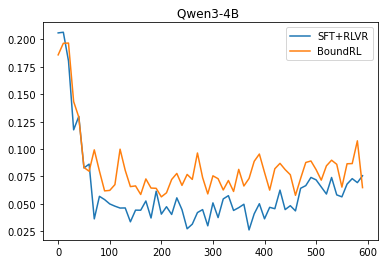}
        \caption{Qwen3-4b}
    \end{subfigure}
    \hfill
    \begin{subfigure}[]{0.3\textwidth}
        \centering
        \includegraphics[width=\textwidth]{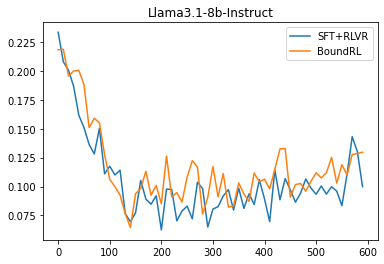}
        \caption{Llama3.1-8b-Instruct}
    \end{subfigure}
    
    \caption{The standard deviation of rewards during training for \task~and SFT+RLVR. Intermediate candidates help \task~mitigate the entropy collapse issue of RLVR.}
    \label{fig:reward_std_curve}
\end{figure*}

In this section, we compare the standard deviation of rewards of SFT+RLVR and \task~during training to evaluate whether \task~can mitigate the entropy collapse issue of RLVR. Specifically, we show the curve of standard deviation of rewards among candidate segmentations generated during rollout along the training. We show the curve of SFT+RLVR and \task~in Fig.\ref{fig:reward_std_curve}. From the figure, we find that the standard deviation of rewards of SFT+RLVR quickly becomes very small, while that of \task~remains stable throughout training. The results show that intermediate candidates help \task~mitigate the entropy collapse issue of RLVR.

\subsection{Qualitative Examples}
\label{app:example}
\begin{figure*}[t]
\centering
\includegraphics[width=0.9\textwidth, keepaspectratio]{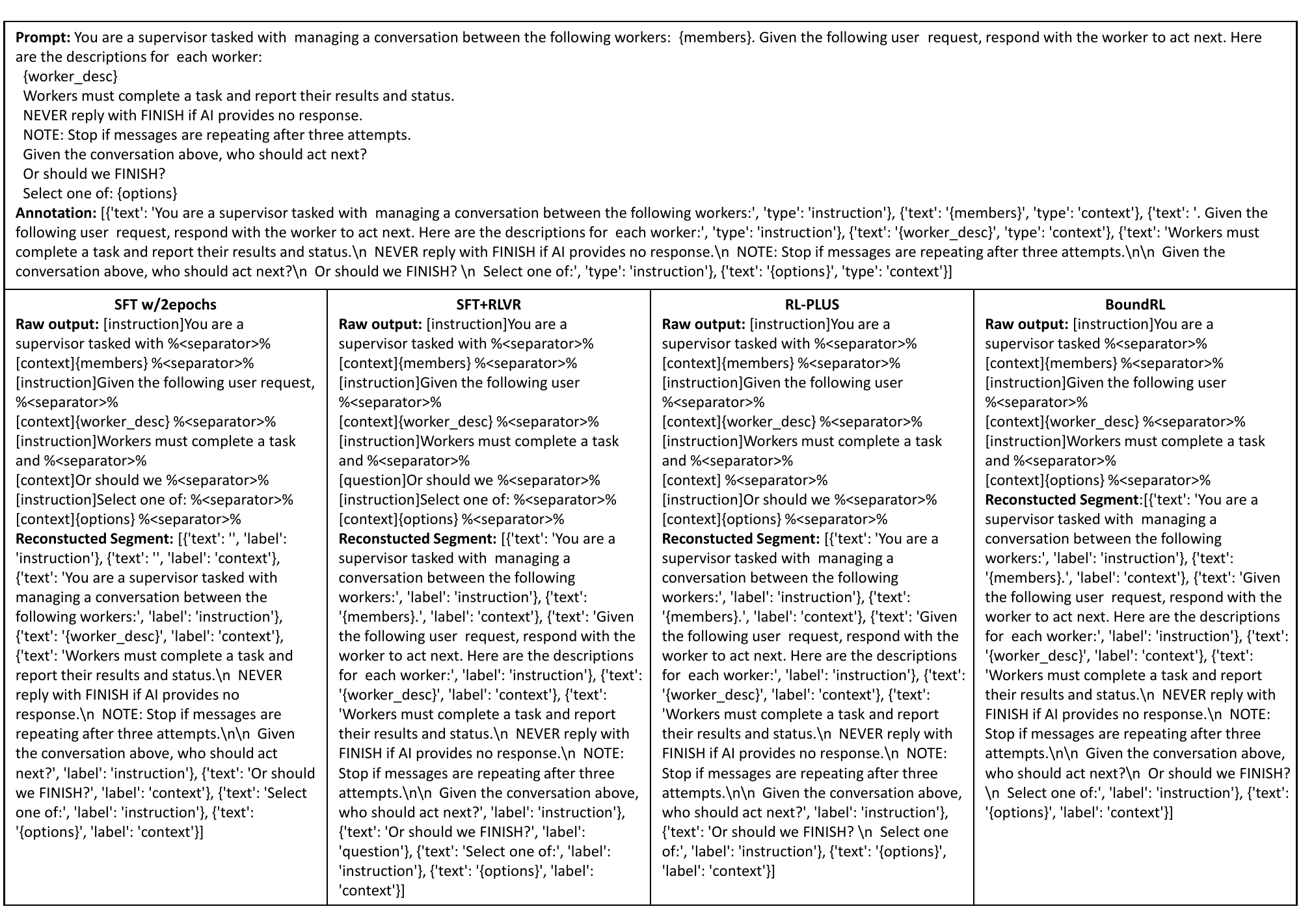}
\caption{Qualitative examples of \task~and other baselines.}
\label{fig:example}
\end{figure*}

In this section, we show qualitative examples of \task~and other baselines. Specifically, we show an example prompt, its corresponding annotated segments, the raw output and reconstructed segments for each method in Fig. \ref{fig:example}. 

\begin{figure*}[t]
\centering
\includegraphics[width=0.9\textwidth, keepaspectratio]{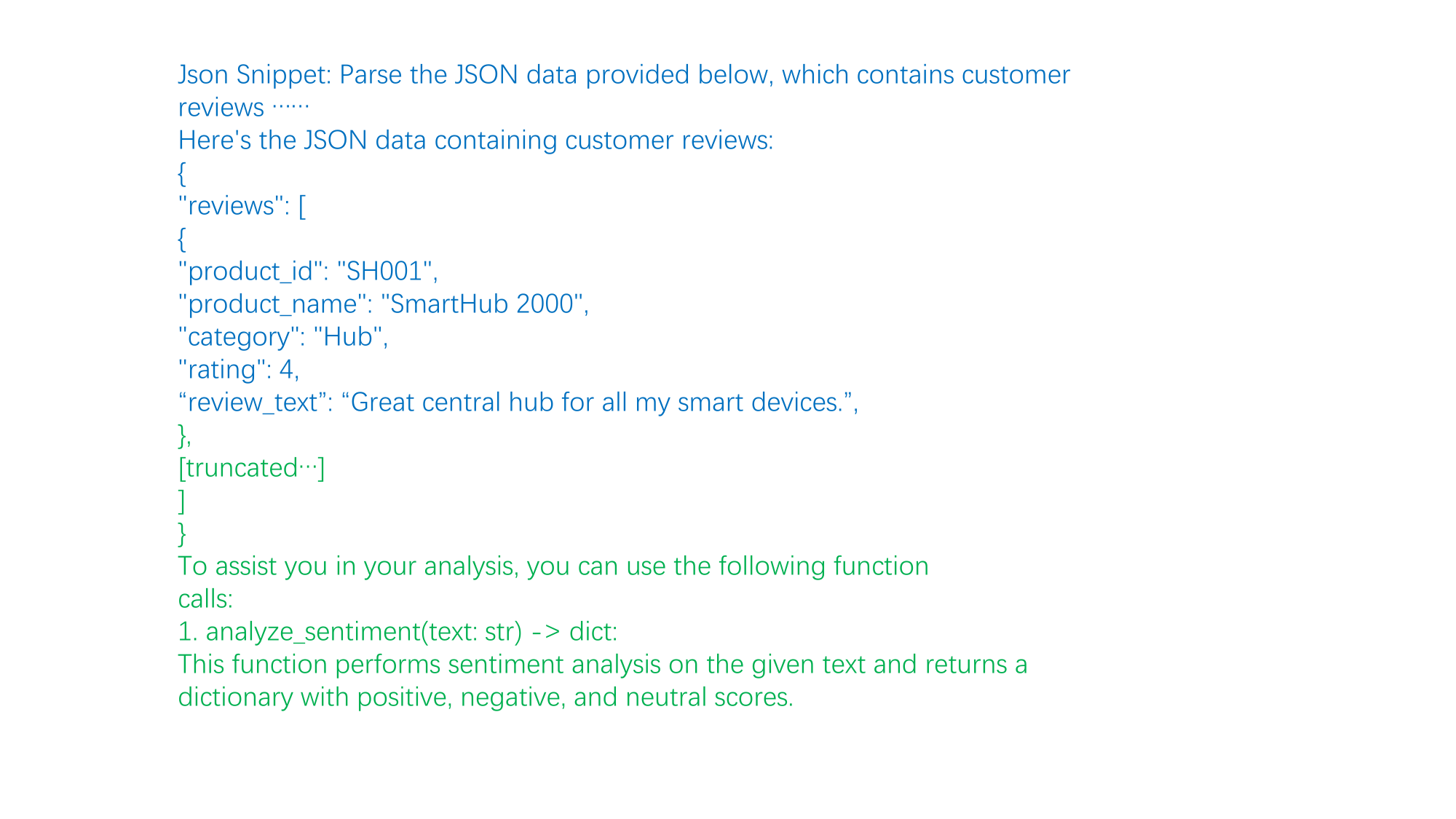}
\caption{Sentences split by the \textrm{Oracle\textsubscript{sent}}'s sentence tokenizer. The first sentence is in \textcolor{blue}{blue}. The second sentence is in \textcolor{green}{green}. The sentence tokenizer cannot effectively handle structured elements that do not follow conventional sentence boundaries.}
\label{fig:sent_example}
\end{figure*}

To analyze the performance of \textrm{Oracle\textsubscript{sent}}, we show sentences split by its sentence tokenizer from a json snippet in Fig. \ref{fig:sent_example}. As shown in the figure, the first half of json snippet and its previous sentence are mistakenly merged into the first sentence. Similarly, the second half of the json snippet and the list of functions are also mistakenly merged into the second sentence. The example shows that the sentence tokenizer cannot effectively handle structured elements that do not follow conventional sentence boundaries. These failures are confirmed quantitatively by our \textrm{Oracle\textsubscript{sent}} in Tab. \ref{tab:RLVR}. Despite having access to ground-truth labels, it achieves only 2.7 and 5.2 EM on Synthetic and Langchain respectively, with high $P_k$ scores (16.5 and 18.9). 

\subsection{Implementation Details of Ablation Studies}
\label{app:ablation}
In this section, we provide more implementation details of the ablated versions of \task. In \task~w/ 2steps, we construct the intermediate candidates by selecting the first perturbation step that has the biggest reward gain over the original candidate segmentation with the medium-level reward. We then select the second perturbation that has the biggest reward gain when the first perturbation step is applied. The intermediate candidate is then constructed by applying the first and second perturbation steps on the original candidate segmentation. For \task~w/o select, we incorporate intermediate candidates for all input texts where the intermediate candidate has a positive reward gain over the original generated candidate segmentation, rather than restricting replacement to the top-$k$ cases. For \task~w/o middle, we construct intermediate candidates by perturbing a randomly sampled candidate segmentation instead of the one with the medium-level reward. Other design choices for these ablated versions are the same as those for \task~for a fair comparison. 

\subsection{Evaluation of \task~on WikiSection}
\label{app:wikisection}
In addition to the \dataset, we also evaluate \task~using Qwen3-1.7b on WikiSection-city dataset. We compare \task~with the same set of baselines as in Sec.\ref{sec:boundrl}. The results are shown in Tab. \ref{tab:wikisection}.

As shown in Tab. \ref{tab:wikisection}, all methods perform almost the same while \task~slightly outperforms all other baselines. The smaller performance gap on WikiSection-city compared with \dataset~is mainly because WikiSection-city is much simpler in terms of structure. Unlike prompts in \dataset, input texts in WikiSection-city only contains the plain text and the ground-truth segmentation is on the sentence level instead of token level. However, even in such cases, \task~still slightly outperforms other baselines.   

\begin{table*}[ht!]
\centering
\resizebox{0.7\textwidth}{!}{
\begin{tabular}{lcccccc}
\toprule
                      &$\rho_{\textrm{rec}}$ & $\textrm{EM}$ & $ P_k$  & $\textrm{F1}_{\textrm{lab}}$ & $\textrm{F1}_{\textrm{char}}$  & Avg.           \\ \midrule
SFT                   & 0.993          & 0.701          & 0.066          & 0.817          & 0.816          & 0.852          \\
SFT w/2epoches        & 0.993          & \textbf{0.713} & 0.064          & 0.819          & 0.825          & 0.857          \\
SFT+RLVR              & 0.994          & 0.711          & 0.063          & \textbf{0.826} & 0.824          & 0.859          \\
SFT+RLVR w/high temp. & \textbf{0.995} & 0.709          & 0.064          & 0.824          & 0.824          & 0.857          \\
RL-PLUS               & 0.993          & 0.709          & 0.063          & 0.819          & 0.825          & 0.857          \\
\task               & 0.994          & 0.712          & \textbf{0.062} & 0.825          & \textbf{0.827} & \textbf{0.859} \\ \bottomrule
\end{tabular}}
\caption{Evaluation of \task~using Qwen3-1.7b on WikiSection-city dataset. The best-performing method highlighted in \textbf{bold}.}
\label{tab:wikisection}
\end{table*}

\end{document}